\ifdefined\XeTeXversion\else\pdfoutput=1\fi 
\documentclass{article}
\usepackage{microtype}
\usepackage{graphicx}
\usepackage{booktabs}
\usepackage{array}
\usepackage{longtable}
\usepackage{float}
\usepackage{needspace}
\usepackage{fancyvrb}
\usepackage{fvextra}
\usepackage{hyperref}

\usepackage[preprint]{wsparxiv}
\usepackage[T1]{fontenc}
\usepackage[utf8]{inputenc}
\usepackage{amsmath}
\usepackage{amssymb}
\usepackage{mathtools}
\usepackage[capitalize,noabbrev]{cleveref}
\icmltitlerunning{What Does Weight-Space Programming Add to Diffusion Editing?}
\begin{document}
\twocolumn[
  \icmltitle{What Does Weight-Space Programming Add to Diffusion Editing?\\ Program Construction versus Score-Guided Refinement}
  \icmlsetsymbol{corr}{$\dagger$}
  \icmlaffilorder{fudan,simis,tencent,nwu}
  \begin{icmlauthorlist}
    \icmlauthor{Zitian Gao}{fudan}
    \icmlauthor{Jingyi Xu}{nwu}
    \icmlauthor{Ziqi Han}{fudan}
    \icmlauthor{Chengkai Lei}{fudan}
    \icmlauthor{Ye Liu}{tencent}
    \icmlauthor{Liangliang Shi}{simis,corr}
  \end{icmlauthorlist}
  \icmlaffiliation{fudan}{Fudan University}
  \icmlaffiliation{simis}{Shanghai Institute for Mathematics and Interdisciplinary Sciences}
  \icmlaffiliation{tencent}{Tencent}
  \icmlaffiliation{nwu}{Northwest University}
  \icmlcorrespondingauthor{Liangliang Shi}{shiliangliang@simis.cn}
  \icmlkeywords{weight-space programming, diffusion model editing, evaluation}
  \vskip 0.3in
]
\printAffiliationsAndNotice{}
\begin{abstract}
Weight-space programming (WSP) lets a language model construct executable weight updates for another network. We study diffusion concept editing through an interface in which programs define training contexts, preservation constraints and update directions, while a shared numerical fitter selects their coefficients. We compare generated programs with published editors, a shared-panel hand-written baseline and budget-matched search over conventional edits on 24 TIMED test edits for Stable Diffusion 1.4. The preregistered rewrite pipeline shows no advantage over the strong baseline. In an exploratory comparison, one-shot Astra programs improve fresh-scene VLM joint success, which requires subject presence together with whole-image target matching, by +10.0 points over the shared-panel baseline, while CLIP relative match shows no detectable change. These programs organise adjustable directions around protected text-state subspaces. Probing and rewriting the Astra programs then raises CLIP by +6.2 points while joint success changes by -9.7. On a fixed 144-image sample, five additional evaluators give point estimates in the same directions under whole-image target matching; checks of subject-local change give less decisive differences. Together, these results identify useful behaviour in complete generated programs and show that score-guided revision need not preserve it: program construction and revision should be evaluated as distinct stages of a WSP pipeline.
\end{abstract}

\section{Introduction}\label{sec:1}

Weight-space programming hands part of model modification to a language model. The LLM reads a network's interface and a task description and writes a program that constructs a weight update \citep{takehi2026sparks,wsp_explore,taghia2026erase,weightloop}. In model editing, the appeal is that the update rule can change with each edit. The LLM can decide which contexts to fit, what to protect and how to organise the update, rather than applying one published algorithm to every edit, and it can revise the program after seeing its effect.

Whether this adds anything is hard to judge in practice. Concept editing already has closed-form editors with public benchmarks, such as TIME \citep{orgad2023time} and UCE \citep{gandikota2024uce}. A WSP pipeline also has components that are not the program. The protection prompts an LLM proposes can be handed to any editor. A numerical fitter chooses how much of each proposed direction to apply. The benchmark score, a CLIP relative match between the edited image and the old and new descriptions, can rise when the image moves toward the target words without the edited subject. Recent WSP studies outside diffusion find that apparent gains can come from such components rather than from what the LLM wrote \citep{wsp_milp}.

We therefore compare LLM-written programs with controls for two of these components and with a check that targets the third. The controls are a hand-written baseline (U1b) that uses the same protection panel and the same fitter, and a random search over the conventional closed-form family with the same candidate-evaluation budget and selection rule. Every unit is scored with CLIP and with a VLM check of subject presence and target-description match, on TIMED's prompts and on fresh-scene prompts written after all units were frozen.

\begin{figure*}[t]
\centering
\includegraphics[width=\textwidth]{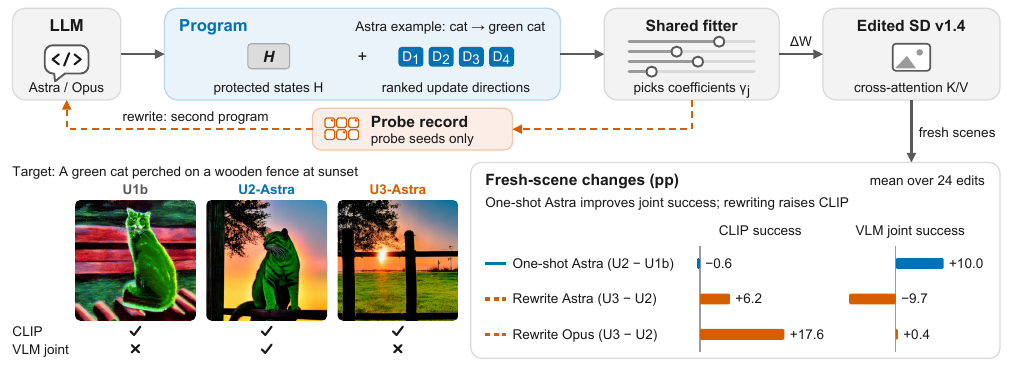}
\caption{Astra's construction gain is lost after score-guided rewriting, even as CLIP rises. For each concept edit, an LLM writes a program that returns ranked update directions for the cross-attention K/V weights of Stable Diffusion v1.4; the one-shot Astra program for cat $\rightarrow$ green cat shown here also protects a set of text states $H$. A shared CMA-ES fitter, with the same budget for every unit, chooses only the coefficients. In the orange dashed loop, the LLM sees a probe record (scores of its directions and fitted updates on probe seeds only) and writes a second program that passes through the same fitter. Bottom right: paired mean changes over the 24 TIMED test edits on fresh-scene prompts written after all units were frozen, in percentage points; one-shot Astra is compared with the shared-panel baseline U1b, each rewrite with the one-shot program of the same LLM. VLM joint success requires the subject to be present and the whole image to match the target description. Bottom left: one illustrative prompt and seed with each unit's recorded CLIP and VLM joint verdicts; in the U3-Astra image CLIP still judges success while the subject is gone. Paired intervals are in Table~\ref{tab:1}.}
\label{fig:1}
\end{figure*}

We compare one-shot program construction under a common fitting protocol, then examine what changes after score-guided rewriting (Fig.~\ref{fig:1}). Under that protocol, \emph{program construction} can add task value. One-shot programs written by Astra raise fresh-scene joint success over the shared-panel baseline by +10.0 [+4.7, +15.4] points, while the CLIP score shows no detectable difference (-0.6 [-7.3, +6.4]). Their code shares one structural pattern: hard projection onto the complement of a protected text-state subspace. Which states go into that subspace, and how the remaining directions are organised, differs from program to program (\S\ref{sec:2}). \emph{Score-guided refinement} does not preserve that value. After probing and rewriting, CLIP generality rises (Opus +17.6, Astra +6.2) and joint success does not: it changes by +0.4 for Opus and -9.7 for Astra. The hand-written extension shows the same direction without any generated program. On fresh scenes, extending and fitting TIME by hand raises CLIP by +26.0 points with no detectable change in joint success, and an LLM-free follow-up on 40 further TIMED edits shows that requiring the subject attenuates a +21.4-point CLIP advantage between conventional editors by 21.7 points.

Our contributions are:
\begin{enumerate}
\item \textbf{An executable WSP pipeline for diffusion concept editing and an analysis of its generated update spaces.} We specify the program--fitter interface and show how a representative generated program combines residual regressions with a shared text-state invariant.
\item \textbf{A controlled comparison of program construction and score-guided revision.} We compare complete generated programs with shared-panel and budget-matched conventional controls on 24 TIMED edits. One-shot Astra programs show an exploratory gain in fresh-scene joint success over the shared-panel baseline, while rewriting raises CLIP without retaining that gain.
\end{enumerate}

\section{Weight-Space Programming for Concept Editing}\label{sec:2}

\subsection{Task, interface and what the LLM is told}\label{sec:2.1}

An edit is a pair \texttt{old $\rightarrow$ new}, such as "a cat" $\rightarrow$ "a green cat". After the edit, prompts that mention the subject in any scene should produce the new attribute, and prompts about other subjects should be unchanged. Programs may modify the key and value projections $W^0_\ell$ of the 16 cross-attention layers of Stable Diffusion 1.4, 32 matrices in total. For each edit, the LLM writes \texttt{build\_directions(ctx)}. The context provides the two prompts, the layer list, the original weights, a text encoder for any prompts the program writes, a tokenizer and TIME's token alignment. The program returns up to eight ranked update directions $D_1, \dots, D_m$, up to four protection prompts and optional initial coefficients.

The prompt states the goal and gives TIME as "a baseline you should try to beat, not the required answer". It includes TIME's objective, its closed form with $\lambda$ = 0.1, its augmentation templates and its reported weaknesses. It also says that a numerical optimiser will choose coefficients for the first four directions by a CLIP score with a protection penalty. The programs are therefore written with the published reference in hand. Appendix~\ref{app:B} reproduces both prompts verbatim.

\subsection{Anatomy of one generated program}\label{sec:2.2}

The one-shot Astra program for "a cat" $\rightarrow$ "a green cat" shows what this interface lets a program build. It writes 15 further source--target context pairs --- scenes, styles, plural cats, mixed cat--dog scenes --- and 30 preservation texts, which supply the regression samples $P^0$ used below and are distinct from the at most four protection prompts returned to the fitter. Using TIME's alignment it forms old-state inputs and target-minus-old residuals, keeping edited-noun samples $(X^0_n, R_n)$ separate from suffix, end-of-text and padding samples $(X^0_c, R_c)$.

It then builds an orthonormal basis $H$ of the empty-prompt states and of the causal-prefix states that the edit leaves unchanged, and with $\Pi = I - HH^\top$ projects its inputs and preservation samples off that span:

\[
X_n = \Pi X^0_n, \qquad X_c = \Pi X^0_c, \qquad P = \Pi P^0 .
\]

One regularised covariance is shared by both sample groups,

\[
C = X_nX_n^\top + X_cX_c^\top + 3PP^\top + \rho I ,
\]

and two residual-transfer operators are solved from it,

\[
A_n = R_nX_n^\top C^{-1}\Pi, \qquad A_c = R^{(32)}_cX_c^\top C^{-1}\Pi ,
\]

where $\rho>0$ is the program's ridge term and $R^{(32)}_c$ is a rank-32 truncation of the context residuals. The implementation uses a Cholesky solve and reapplies the projection afterwards. Each operator becomes a weight direction through $W^0_\ell A$, with key and value matrices kept apart, giving four ranked directions: noun values, context values, noun keys, context keys.

Two properties follow. First, the shared covariance ties the two components together while the separate right-hand sides let the fitter move them independently. Second, because $A_nH = A_cH = 0$, every returned direction satisfies $D_{j,\ell}H = 0$, so any combination the fitter chooses obeys

\[
\Big(\textstyle\sum_j \gamma_j D_{j,\ell}\Big)H = 0
\]

up to numerical precision; normalisation and orthogonalisation preserve this because they take linear combinations. The construction therefore hands the fitter adjustable components that share one linear invariant, and the invariant concerns token states; the image-level check below is a separate measurement. Box~\hyperref[box:1]{1} (Appendix~\ref{app:B}) summarises the division of labour for this program.

This is a representative program, not a template the LLM was given. Others organise their directions differently: the bed $\rightarrow$ bunk-bed program returns full value transfer, full key transfer, a subject-only value component, and a component restricted to selected low-resolution layers. We therefore report recurring structure separately from this example's particular ranks and ordering.

\subsection{What recurs across the programs}\label{sec:2.3}

Coding all 96 final programs from their syntax trees (Table~\ref{tab:A5}) shows one structural regularity that separates the two models. Every one-shot Astra program builds an SVD basis from text states and projects its update off it (24/24), and all of them encode the empty prompt among the states so protected (24/24). No one-shot Opus program does either (0/24 and 0/24): they are TIME-style ridge regressions with soft protection terms, often returned at several regularisation strengths or layer groups. What goes into the protected set is not uniform: the cat program also fixes the unchanged causal prefixes of its training pairs, the peonies program fixes the empty prompt together with four explicit anchor prompts, and some programs add further texts. The noun/context split described in \S\ref{sec:2.2} is the cat program's own organisation. These codes come from source structure; they are not a runtime verification of every program's invariance. Null-space preservation has precedents in diffusion editing, including SPEED \citep{li2026speed} and ACE \citep{wang2025ace}. Here we characterise how the generated programs combine it with edit-specific contexts and adjustable update directions.

\subsection{Fitting and selecting an update}\label{sec:2.4}

The fitter walks the returned directions in rank order, skips zero or nearly dependent ones, and keeps at most four independent components by two-pass Gram--Schmidt. Flattening all editable matrices,

\[
\operatorname{vec}\,\Delta W(\beta) = Q\beta, \qquad Q^\top Q = I .
\]

CMA-ES evaluates 24 coefficient candidates inside a Frobenius ball of radius three times the norm of TIME's update for that edit. Each candidate renders the old prompt with four fixed probe seeds and each protection prompt with the first two of them, and is scored by

\[
J = E - 5\max(0,\, D - D_{\mathrm{TIME}}) - 5\max(0,\, S_{\mathrm{ref}} - S).
\]

$E$ is the gain in the CLIP margin between the target and old descriptions relative to the unedited model. $D$ is the mean CLIP dissimilarity between edited and unedited renderings of the protection panel, and $D_{\mathrm{TIME}}$ is TIME's value. $S$ is CLIP similarity to "a photo of $\langle$subject$\rangle$", with $S_{\mathrm{ref}}$ measured on the unedited model drawing the target prompt. Among candidates satisfying $D \le D_{\mathrm{TIME}} + 0.01$ and $S \ge S_{\mathrm{ref}} - 0.01$ the pipeline takes the largest $E$; only if none qualifies does it take the largest $J$. The pipeline therefore already contains a CLIP proxy for subject retention. Fitting uses CLIP ViT-L/14 while the reported evaluation uses ViT-bigG-14. Neither TIMED's generality prompts nor the fresh-scene prompts enter fitting.

The program's text-state invariant, the fitter's image-based penalties and the VLM evaluation answer three different questions. \S\ref{sec:3} defines the image-level endpoint.

\textbf{Probe and rewrite (U3).} After the one-shot programs (U2) are fitted, the LLM receives a probing record. It contains contact sheets of a library of directions at three doses (the library includes the U2 programs' own directions and TIME/U1b group directions) and of every unit's fitted update, with their $E$, $D$ and $S$ values, a median of 912 images per edit. The rewrite prompt states the subject tolerance explicitly. The LLM then writes a second program, which goes through the same fitter. U3 is a pipeline comparison: it differs from U2 in information, generation and fitting, and we compare the pipeline as a whole.

\begin{figure*}[t]
\centering
\includegraphics[width=\textwidth]{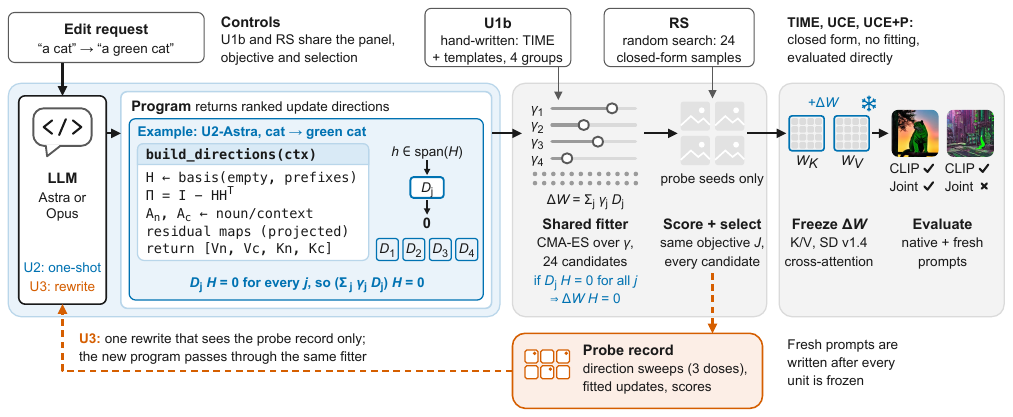}
\caption{Program construction, shared selection and evaluation. The blue inset shows the U2-Astra cat program: it builds an orthonormal basis $H$ of empty-prompt and unchanged-prefix states and four directions whose input side annihilates it, $D_jH = 0$, so every combination the fitter can choose keeps the invariant. The shared CMA-ES fitter keeps at most four independent directions and evaluates 24 coefficient candidates; U1b directions enter the same fitter, RS samples 24 closed-form candidates directly, and TIME, UCE and UCE+P are closed-form updates evaluated directly. Every candidate is scored by the same objective on probe seeds only. For U3 (dashed), a probe record of direction sweeps and fitted updates is returned to the LLM, which writes one rewritten program that passes through the same fitter. Selected updates are frozen before evaluation, and fresh prompts are written after all units are frozen. The 24-candidate budget applies to this stage only. Thumbnails are illustrative fresh-scene samples of the U2-Astra cat edit with their recorded CLIP and VLM joint verdicts; probe-seed tiles are schematic.}
\label{fig:2}
\end{figure*}

\section{Controlled Comparison}\label{sec:3}

\textbf{Units.}

\begin{table*}[t]
\centering\scriptsize\setlength{\tabcolsep}{3pt}
\caption{Units in the main comparison. All fitted units share the protection panel, the candidate-evaluation budget, the fitting score and the selection rule; RS draws conventional closed-form candidates directly instead of searching coefficients of returned directions.}
\label{tab:inline1}
\begin{tabular}{p{0.17\linewidth}p{0.25\linewidth}p{0.25\linewidth}p{0.25\linewidth}}
\toprule
unit & update directions & coefficients & shares with programs \\
\midrule
TIME, UCE & published closed form & fixed & --- \\
UCE+P & UCE with the shared panel as preserved concepts & fixed & protection panel \\
U1b (hand-written) & TIME + 6 further templates + preservation term on the shared panel, split into 4 layer groups & CMA-ES, 24 evaluations & panel, fitter, objective, selection \\
RS (random search) & regulariser, protection weight, strength, template/panel/layer-group subsets sampled from the conventional closed-form family & none: each sample is one candidate, 24 candidates & panel, objective, selection, probe seeds \\
U2 (one-shot LLM) & program & CMA-ES, 24 evaluations & --- \\
U3 (probe + rewrite) & second program, after the probing record & CMA-ES, 24 evaluations & --- \\
\bottomrule
\end{tabular}
\end{table*}

The panel shared by UCE+P, U1b and RS contains prompts that the LLMs proposed, so U1b is a shared-panel baseline rather than an LLM-free one. It does not receive the contexts that a program writes for itself (a median of 96 encoded texts per U2-Astra program). RS samples conventional closed-form edits with a single strength over the chosen layer groups, whereas U1b fits four group coefficients, so RS does not contain fitted U1b updates. The LLMs are Astra (gpt-6-astra) and Opus (Claude Opus 5.5).

\textbf{Edits and prompts.} Before the main comparison, we drew 24 of TIMED's 104 test edits uniformly at random with a fixed seed. TIME had been run on all 104 before, so the draw is a frozen prospective comparison rather than a held-out set. Native prompts are TIMED's own: the old prompt for efficacy and five positives for generality. After all units were frozen, Gemini wrote four fresh-scene positives and eight near-neighbour negatives per edit. Any text that any program had encoded was excluded, and Opus checked every pair (Appendix~\ref{app:C}). Each prompt is rendered with 12 seeds.

\textbf{Metrics.} We use two kinds of score on two prompt populations:

\begin{table}[t]
\centering\scriptsize\setlength{\tabcolsep}{3pt}
\caption{The two kinds of score on the two prompt populations.}
\label{tab:inline2}
\begin{tabular}{p{0.17\linewidth}p{0.37\linewidth}p{0.37\linewidth}}
\toprule
 & CLIP relative match (ViT-bigG-14, TIMED convention) & VLM joint (Qwen3.8-27B: subject present $\wedge$ matches target description) \\
\midrule
native TIMED prompts & efficacy, generality, specificity & joint efficacy, joint generality \\
fresh-scene prompts & generality; specificity on fresh negatives & fresh-scene joint success; subject presence \\
\bottomrule
\end{tabular}
\end{table}

CLIP counts an image as a success when it is closer to the edited description than to the unedited one. Efficacy uses the old prompt, generality the positives, and specificity the negatives, where the unedited description must win. The VLM judge answers two yes/no questions per image. The first asks whether the subject is present; it may be recoloured or restyled, but it must not be a different kind of object. The second asks whether the image matches the target description. A joint success needs both "yes" probabilities to be at least 0.5. The judge was selected on a separate set of 274 human-labelled questions against a pre-set bar of macro balanced accuracy $\geq$ 0.90 (Qwen3.8-27B: 0.955). These were generic questions about named subjects; the two questions above were adopted later and were not re-run on that set (Appendix~\ref{app:D}, Table~\ref{tab:A2}).

\textbf{Additional evaluators.} We also re-evaluate a fixed 144-image sample with five models spanning four providers, asking each question in a separate call. Both endpoints there require a positive subject-presence judgement: \texttt{full} additionally asks whether the image matches the complete target description, as above, and \texttt{local} instead asks whether the named change occurs on the subject itself. \texttt{full} does not check attribute binding item by item, and \texttt{local} does not require the scene to match. Table~\ref{tab:3} reports both and records each model's role in the workflow.

\textbf{Statistics.} Intervals are paired percentile bootstraps over edits (B = 50,000). Among the endpoints reported here, the preregistered contrasts for the main comparison are U3-Opus against U1b, TIME, UCE, UCE+P and U2-Opus, and U1b against TIME, on native joint efficacy (J-eff), native joint generality (J-gen), fresh-scene joint success (F-joint) and fresh-negative specificity (F-spec). Other contrasts in this matrix, including every comparison on subject presence, are exploratory. We also report an exploratory analysis of all 288 unit-pair $\times$ endpoint contrasts, with paired sign-flip randomisation p-values and Holm and Benjamini--Hochberg adjustments within each endpoint and over all cells (Table~\ref{tab:A6}).

\section{What Does Program Construction Add?}\label{sec:4}

\begin{table*}[t]
\centering\footnotesize\setlength{\tabcolsep}{4pt}
\caption{Controlled comparison on 24 pre-drawn TIMED test edits. \textbf{A}: unit means (\%). \textbf{B}: paired differences in percentage points with 95\% paired bootstrap intervals over edits (50{,}000 resamples, unadjusted). Bold: the rounded interval excludes zero. $\star$: unit pair in the preregistration for the main comparison, whose preregistered endpoints are J-eff, J-gen, F-joint and F-spec; F-subj and F-CLIP are exploratory for every pair. J: native joint (VLM subject present $\wedge$ matches target description); F-joint: fresh-scene joint success; F-subj: fresh-scene subject presence; F-CLIP: fresh-scene CLIP relative-match generality; F-spec: CLIP specificity on fresh negatives. LLM costs are in Table~\ref{tab:A7}; all 288 unit-pair contrasts are in Table~\ref{tab:A6}.}
\label{tab:1}
\begin{tabular}{lcccccc}
\toprule
& \multicolumn{2}{c}{Native prompts} & \multicolumn{3}{c}{Fresh positives} & \multicolumn{1}{c}{Fresh neg.} \\
\cmidrule(lr){2-3} \cmidrule(lr){4-6} \cmidrule(lr){7-7}
& J-eff & J-gen & F-joint & F-subj & F-CLIP & F-spec \\
\midrule
\multicolumn{7}{l}{\textbf{A}\quad Unit means (\%)} \\
\addlinespace[1pt]
TIME & 71.2 & 38.1 & 28.0 & 72.1 & 55.1 & 71.7 \\
UCE & 40.6 & 24.5 & 21.5 & 69.6 & 46.9 & 73.0 \\
UCE+P & 39.6 & 23.8 & 20.8 & 71.5 & 40.3 & 86.8 \\
\addlinespace[2pt]
U1b & 74.7 & 48.3 & 27.7 & 64.0 & 81.2 & 78.5 \\
RS & 75.0 & 43.5 & 32.0 & 70.0 & 71.4 & 77.1 \\
\addlinespace[2pt]
U2-Astra & 73.6 & 49.4 & 37.7 & 68.8 & 80.6 & 84.5 \\
U2-Opus & 59.7 & 39.4 & 30.4 & 70.5 & 63.3 & 85.8 \\
\addlinespace[2pt]
U3-Astra & 73.3 & 44.8 & 28.0 & 60.8 & 86.8 & 78.2 \\
U3-Opus & 74.3 & 42.4 & 30.8 & 61.9 & 80.9 & 80.3 \\
\addlinespace[2pt]
target ref. & 78.8 & 57.1 & 40.4 & 75.3 & 79.6 & -- \\
\midrule
\multicolumn{7}{l}{\textbf{B}\quad Paired differences (pp)} \\
\addlinespace[1pt]
U3-Opus $-$ U1b\,$^\star$ & $-$0.3 & \textbf{$-$6.0} & +3.1 & $-$2.1 & $-$0.3 & +1.7 \\
 & {\scriptsize[$-$6.2, +5.9]} & {\scriptsize[$-$10.3, $-$1.6]} & {\scriptsize[$-$3.0, +8.6]} & {\scriptsize[$-$10.4, +5.7]} & {\scriptsize[$-$8.2, +6.5]} & {\scriptsize[$-$2.4, +5.8]} \\[1.5pt]
U3-Opus $-$ TIME\,$^\star$ & +3.1 & +4.2 & +2.9 & \textbf{$-$10.2} & \textbf{+25.8} & \textbf{+8.6} \\
 & {\scriptsize[$-$7.3, +13.2]} & {\scriptsize[$-$4.4, +13.3]} & {\scriptsize[$-$4.1, +9.8]} & {\scriptsize[$-$17.0, $-$4.1]} & {\scriptsize[+13.5, +37.8]} & {\scriptsize[+3.3, +13.8]} \\[1.5pt]
U3-Opus $-$ RS & $-$0.7 & $-$1.2 & $-$1.2 & \textbf{$-$8.1} & \textbf{+9.5} & +3.1 \\
 & {\scriptsize[$-$9.0, +8.0]} & {\scriptsize[$-$8.4, +6.4]} & {\scriptsize[$-$8.2, +4.8]} & {\scriptsize[$-$15.6, $-$0.7]} & {\scriptsize[+1.1, +18.8]} & {\scriptsize[$-$2.0, +7.9]} \\[1.5pt]
\addlinespace[2pt]
U2-Astra $-$ U1b & $-$1.0 & +1.0 & \textbf{+10.0} & +4.8 & $-$0.6 & \textbf{+5.9} \\
 & {\scriptsize[$-$10.4, +7.3]} & {\scriptsize[$-$4.0, +6.1]} & {\scriptsize[+4.7, +15.4]} & {\scriptsize[$-$4.8, +14.4]} & {\scriptsize[$-$7.3, +6.4]} & {\scriptsize[+2.5, +9.3]} \\[1.5pt]
U2-Opus $-$ U1b & \textbf{$-$14.9} & \textbf{$-$9.0} & +2.7 & +6.5 & \textbf{$-$17.9} & \textbf{+7.2} \\
 & {\scriptsize[$-$30.9, $-$0.7]} & {\scriptsize[$-$18.4, $-$0.1]} & {\scriptsize[$-$4.9, +9.5]} & {\scriptsize[$-$1.6, +15.1]} & {\scriptsize[$-$34.0, $-$3.1]} & {\scriptsize[+3.7, +11.1]} \\[1.5pt]
U2-Astra $-$ RS & $-$1.4 & \textbf{+5.8} & +5.6 & $-$1.2 & \textbf{+9.2} & \textbf{+7.3} \\
 & {\scriptsize[$-$11.1, +7.6]} & {\scriptsize[+0.6, +11.0]} & {\scriptsize[0.0, +11.5]} & {\scriptsize[$-$8.7, +6.4]} & {\scriptsize[+0.8, +18.1]} & {\scriptsize[+2.7, +12.0]} \\[1.5pt]
\addlinespace[2pt]
U3-Opus $-$ U2-Opus\,$^\star$ & +14.6 & +3.0 & +0.4 & \textbf{$-$8.6} & \textbf{+17.6} & $-$5.5 \\
 & {\scriptsize[$-$0.7, +31.9]} & {\scriptsize[$-$6.4, +12.8]} & {\scriptsize[$-$6.8, +7.5]} & {\scriptsize[$-$16.1, $-$1.6]} & {\scriptsize[+5.4, +31.9]} & {\scriptsize[$-$11.3, 0.0]} \\[1.5pt]
U3-Astra $-$ U2-Astra & $-$0.3 & $-$4.6 & \textbf{$-$9.7} & $-$8.0 & \textbf{+6.2} & \textbf{$-$6.2} \\
 & {\scriptsize[$-$9.4, +8.0]} & {\scriptsize[$-$9.9, +0.8]} & {\scriptsize[$-$16.6, $-$2.9]} & {\scriptsize[$-$17.2, +1.2]} & {\scriptsize[+1.3, +11.6]} & {\scriptsize[$-$11.5, $-$1.6]} \\[1.5pt]
\addlinespace[2pt]
U1b $-$ TIME\,$^\star$ & +3.5 & \textbf{+10.2} & $-$0.3 & \textbf{$-$8.2} & \textbf{+26.0} & \textbf{+6.9} \\
 & {\scriptsize[$-$6.2, +12.8]} & {\scriptsize[+3.3, +17.5]} & {\scriptsize[$-$5.7, +5.3]} & {\scriptsize[$-$16.1, $-$1.0]} & {\scriptsize[+17.1, +36.3]} & {\scriptsize[+2.3, +11.8]} \\[1.5pt]
\bottomrule
\end{tabular}
\end{table*}

All four LLM pipelines produced an executed update on all 24 edits. First attempts succeeded on 24/24 (U2-Astra), 20/24 (U2-Opus), 21/24 (U3-Astra) and 23/24 (U3-Opus) edits, and one repair round fixed the rest. On 4/24 edits for U2-Opus and 1/24 for U3-Opus, the fitter selected a zero update; these count as produced by the pipeline.

\textbf{The preregistered primary method shows no advantage over the strong controls.} U3-Opus $-$ U1b is -0.3 [-6.2, +5.9] on native joint efficacy and -6.0 [-10.3, -1.6] on native joint generality. Against TIME it is +3.1 [-7.3, +13.2] on native joint efficacy, and against random search, which was not in the preregistered list, -0.7 [-9.0, +8.0]. On native joint efficacy, U2-Astra (73.6), U3-Opus (74.3), RS (75.0) and U1b (74.7) are close, well above UCE (40.6). U2-Opus is lower (59.7).

\textbf{One-shot Astra programs add fresh-scene joint success.} What is measured here is the whole program package: the update geometry the program builds, together with the contexts it writes for itself. U2-Astra $-$ U1b is +10.0 [+4.7, +15.4] on fresh-scene joint success, with no detectable difference in CLIP generality (-0.6 [-7.3, +6.4]; Fig.~\ref{fig:1}). In absolute terms, U2-Astra reaches 37.7\% against 27.7\% for U1b and 40.4\% for the unedited model drawing the target prompt. The gain is a per-edit pattern: it is positive on 18 edits and negative on 4. It is exploratory: BH-adjusted across all 288 contrasts p = 0.007, Holm within its endpoint 0.051 (at the boundary), and it does not survive Holm over all cells (0.343).

\textbf{Fresh-scene effects across evaluators.} On a fixed 144-image sample (two fresh-scene slots per edit shared by the three units; Appendix~\ref{app:H}), all five additional evaluators give a positive \texttt{full} estimate for U2-Astra versus U1b and a negative estimate for U3-Astra versus U2-Astra. Under \texttt{local}, four construction estimates are smaller and GPT's is unchanged; all \texttt{local} intervals include zero. The panel supports the \texttt{full} direction on this sample, while local attribute gains remain uncertain. An earlier GPT evaluation that asked both questions in one call, on a six-seed sample of all units, gave smaller construction and rewrite contrasts (+1.9 and -1.9 points). Appendix~\ref{app:G} reports the protocol differences and the shared-image comparison, in which separate question calls did not restore the construction gain.

\begin{table*}[t]
\caption{Five additional evaluators on one fixed sample of 144 images (24 edits, two paired image slots per edit,
three units), each question asked in a separate call. Cells: paired difference (percentage points) and 95\% edit-bootstrap
interval, one shared resample index per column; bold: interval excludes zero, unadjusted. Both target questions require a
positive subject-presence judgement: \emph{full} also asks for a whole-image match to the target description (the endpoint
used throughout), \emph{local} for the named change on the subject itself. Unsure answers stay in the denominator as
unsuccessful. --: local question not evaluated for the primary judge. The last row is the primary judge on the full evaluation sample of the main comparison, a different image population.
Provenance: GPT 6.1 and Qwen3 VL share providers with the Astra programmer and the primary judge; Claude Opus 5.5 wrote the
separate Opus arms, which are not in this table, and checked fresh prompts; the Gemini models share a provider with the
fresh-prompt generator.}
\label{tab:3}
\centering
\small
\setlength{\tabcolsep}{5pt}
\begin{tabular}{lcccc}
\toprule
& \multicolumn{2}{c}{Construction: U2-Astra $-$ U1b} & \multicolumn{2}{c}{Revision: U3-Astra $-$ U2-Astra} \\
\cmidrule(lr){2-3} \cmidrule(lr){4-5}
Evaluator & Full & Local & Full & Local \\
\midrule
Gemini 4 carbon & \textbf{+14.6\,[+2.1, +27.1]} & +4.2\,[$-$14.6, +20.8] & \textbf{$-$16.7\,[$-$31.2, $-$2.1]} & $-$8.3\,[$-$25.0, +10.4] \\
Gemini 3.1 pro & +16.7\,[0.0, +33.3] & +6.2\,[$-$8.3, +20.8] & \textbf{$-$18.8\,[$-$33.3, $-$4.2]} & $-$8.3\,[$-$25.0, +8.3] \\
GPT 6.1 sol & +12.5\,[0.0, +25.0] & +12.5\,[$-$2.1, +29.2] & \textbf{$-$22.9\,[$-$39.6, $-$6.2]} & 0.0\,[$-$14.6, +16.7] \\
Claude Opus 5.5 & +10.4\,[$-$4.2, +25.0] & +4.2\,[$-$12.5, +18.8] & $-$2.1\,[$-$16.7, +12.5] & +4.2\,[$-$10.4, +20.8] \\
Qwen3 VL 235B & +10.4\,[$-$2.1, +22.9] & +2.1\,[$-$4.2, +8.3] & \textbf{$-$18.8\,[$-$35.4, $-$2.1]} & +2.1\,[$-$8.3, +14.6] \\
\midrule
Primary judge, same sample & \textbf{+20.8\,[+4.2, +37.5]} & -- & \textbf{$-$20.8\,[$-$37.5, $-$4.2]} & -- \\
\midrule
Primary judge, full evaluation sample & \textbf{+10.0\,[+4.7, +15.4]} & -- & \textbf{$-$9.7\,[$-$16.6, $-$2.9]} & -- \\
\bottomrule
\end{tabular}
\end{table*}

The clearest signal is for Astra on fresh scenes. On TIMED's native positives, the same contrast is +1.0 [-4.0, +6.1]. For one-shot Opus programs it is +2.7 [-4.9, +9.5]. Against random search, the strongest conventional control on this endpoint, U2-Astra's difference is +5.6 [0.0, +11.5], an interval whose lower end is zero. Against TIME it is +9.7 [+3.6, +15.7].

\textbf{Where the gain sits.} Table~\ref{tab:2} splits the contrast by subject type and by judge question. On the 18 object edits, joint success rises by +7.2 [+1.7, +12.7]. There the difference sits in the target-description question (+6.7 [+2.1, +11.3]), with no difference in subject presence (+0.2 [-9.6, +9.0]). The 6 person and brand edits show a larger gain in both questions (joint +18.4 [+7.3, +29.5]) and contribute 46\% of the summed per-edit gain. The result does not depend on the judge's cut-off: from 0.3 to 0.9, and with a soft score, every interval excludes zero (+5.6 [+1.6, +9.3] at 0.9; Table~\ref{tab:2}). The gain persists without literal scene-word overlap (Appendix~\ref{app:A}).

\textbf{Specificity has to be read with efficacy.} On fresh negatives U2-Astra is more specific than U1b (+5.9 [+2.5, +9.3]) at similar CLIP generality; U2-Opus is too (+7.2 [+3.7, +11.1]) but edits less, its CLIP generality lower by -17.9 points.

\begin{figure}[t]
\centering
\includegraphics[width=\linewidth]{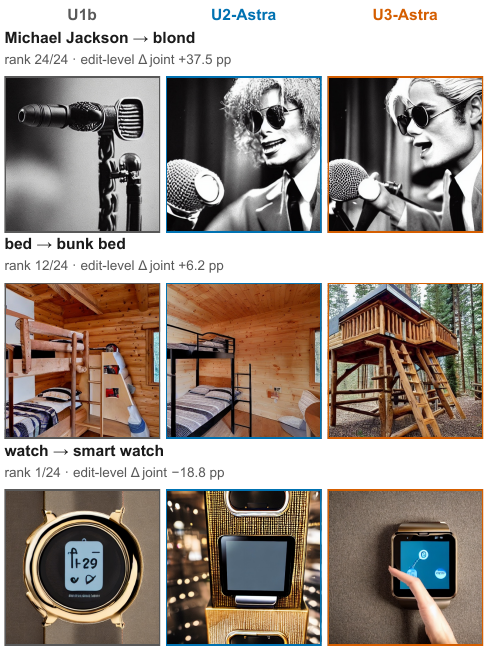}
\caption{Fixed-rule examples across the construction effect. Rows are ranks 24, 12 and 1 of the 24 edits ordered by the U2-Astra $-$ U1b gain in fresh-scene joint success, the middle row being the lower of the two central values (+6.2, +10.4). The gain printed above each row is the edit-level average over all of that edit's fresh-scene images, not a score for the displayed image. Each row uses one prompt and seed across the three units, taken from the frozen audit sample: "Michael Jackson singing into a vintage microphone in a recording studio", "A bed inside a cozy cabin" and "A watch resting on a velvet pillow inside a display case".}
\label{fig:4}
\end{figure}

\section{Does Score-Guided Refinement Preserve the Gain?}\label{sec:5}

\textbf{CLIP gains do not reliably track task gains.} For Astra, rewriting raises fresh-scene CLIP relative match by +6.2 [+1.3, +11.6] while changing joint success by -9.7 [-16.6, -2.9]. All five additional evaluators give a negative \texttt{full} rewrite estimate on the audit sample, four with intervals excluding zero. The Opus rewrite and the hand-written extension also raise CLIP, while their primary-judge joint-success differences remain inconclusive. These units therefore improve on the selection proxy without consistent improvement on the task check.

For Opus, rewriting changes CLIP generality by +17.6 [+5.4, +31.9]; its joint success changes by +0.4 [-6.8, +7.5] (Fig.~\ref{fig:1}). Subject presence changes by -8.6 [-16.1, -1.6] (Opus) and -8.0 [-17.2, +1.2] (Astra; F-subj in Table~\ref{tab:1}). Fresh-negative specificity also falls: -5.5 [-11.3, 0.0] and -6.2 [-11.5, -1.6]. The Opus unit pair was preregistered on J-eff, J-gen, F-joint and F-spec; F-subj and F-CLIP, and the Astra arm, are exploratory. Under Benjamini--Hochberg over all cells, the Astra joint-success decline has adjusted p 0.036 and the Opus CLIP rise 0.033. On native joint efficacy, the Opus rewrite's point estimate is higher (+14.6 [-0.7, +31.9]), largely because four U2-Opus edits with zero updates receive non-zero updates. Without those four edits, the fresh-scene pattern remains (sensitivity analysis): CLIP +10.4, joint -3.1, subject -8.5 [-17.3, -0.5].

\textbf{Access to the one-shot directions did not preserve the gain.} Source-code inspection finds references to U2-Astra library directions in 21/24 Astra rewrites and 19/24 Opus rewrites. Several rewrites also add an explicit subject-identity direction, a response to the subject loss visible on the probe sheets. Rewrites do more than recombine the library: the median share of the final U3-Astra update outside the library's span is 0.51. After fitting and selection, U3-Astra's fresh joint success is close to the hand-written baseline (U3-Astra $-$ U1b +0.3 [-6.4, +6.8]). A code reference does not show that a direction was returned among the four fitted ones or kept a non-zero weight, and our design does not separate the rewrite from the CLIP-guided fitting that follows it.

\textbf{The same direction in a hand-written extension.} U1b is TIME extended with six further augmentation templates, a preservation term on the shared panel, a four-group split and CLIP-guided fitting; its panel contains LLM-proposed prompts, and it contains no generated program. Relative to TIME it also lowers fresh subject presence by -8.2 [-16.1, -1.0], while its native joint generality is higher (+10.2 [+3.3, +17.5]). The extension bundles templates, panel, grouping and fitting together, so these outcomes belong to the extension as a whole and not to fitting alone. Across the nine evaluated units, fresh CLIP generality and fresh subject presence are negatively associated (Pearson r = -0.78; descriptive). The direction matches the rewrite result even though the fitting objective already contains a CLIP subject term (\S\ref{sec:2.4}).

\textbf{A conventional-editor follow-up.} In a preregistered follow-up, we compared a rule-based extension of TIME with TIME on the 23 object edits among 40 further TIMED test edits outside the 24-edit main comparison. The extension uses U1b's regularised TIME construction, grouping, fitter, budget and selection rule, with a protection panel generated by fixed rules only. Its CLIP generality advantage, +21.4 [+14.9, +28.6], is attenuated by 21.7 [13.5, 30.6] once subject presence is required, and all 7/7 preregistered contrasts pass Holm (Table~\ref{tab:A3}). This endpoint combines CLIP success with a subject check rather than the target-matching question used in the main comparison. Two raters also found subject-absent images among CLIP successes (weighted 19.86\% and 18.25\%); Appendix~\ref{app:F} reports their labels, agreement and method contrasts.

\begin{figure}[t]
\centering
\includegraphics[width=\linewidth]{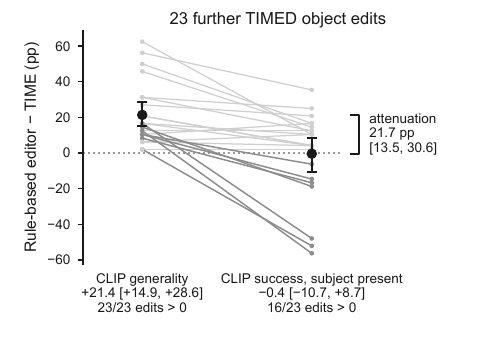}
\caption{Conventional-editor follow-up on 23 further TIMED object edits: the rule-based editor's per-edit CLIP-generality advantage over TIME, before and after requiring subject presence. Each line is one edit, darker where the subject-filtered difference is zero or below; black: mean and 95\% paired bootstrap interval, printed under each column.}
\label{fig:3}
\end{figure}

\section{Related Work}\label{sec:6}

\textbf{Weight-space programming and program generation from feedback.} LLMs have written weight edits for reasoning models \citep{takehi2026sparks,wsp_explore}, for robot policies \citep{wsp_robot}, for scoring networks in solvers \citep{wsp_milp}, and as fixed reusable editors that update predictor weights from experimental feedback \citep{weightloop}. Gains can come from cheap rules rather than from what the LLM wrote \citep{wsp_milp}, and successful fixes are often precise and local \citep{wsp_robot}. \citet{taghia2026erase} has an LLM edit diffusion weights directly to erase a style, with UCE, ESD and budget-matched random search as controls, and reports a benefit from feedback. More broadly, WSP belongs to a line in which an LLM proposes programs or prompts and an evaluator guides their selection \citep{funsearch2024,ma2024eureka,yang2024opro}. Here a CLIP proxy guides the search, so we evaluate the programs both by that score and by a separate task check, to see what construction and score-guided rewriting contribute in targeted attribute edits where the subject must survive.

\textbf{Diffusion editing and constrained updates.} TIME \citep{orgad2023time} and UCE \citep{gandikota2024uce} edit cross-attention projections in closed form, and ReFACT \citep{arad2024refact} edits the text encoder. Null-space preservation has been studied both for language-model editing \citep{fang2025alphaedit} and for diffusion concept erasure: SPEED \citep{li2026speed} constrains start-of-text and null-text states explicitly, and ACE \citep{wang2025ace} applies null-space constraints to cross-attention updates.

\textbf{Evaluation of concept edits and proxy feedback.} Relative CLIP preference for a target over a source description can count inadequate outputs as successful edits. CAKE \citep{gu2024cake} replaces this rule with prompt-specific absolute CLIP thresholds calibrated on unedited target-prompt generations and validates the criterion against VLM and human labels; EMCID \citep{xiong2024emcid} measures source forgetting and destination acquisition through changes in ImageNet-classifier confidence; and for erasure, EraseBench \citep{erasebench} and HUB \citep{moon2025hub} use object- or entity-level visual questions to measure collateral loss of non-target content. CLIP-style scores have documented weaknesses in attribute binding and compositional matching \citep{yuksekgonul2023bow,thrush2022winoground}; TIFA and DSG decompose text--image evaluation into questions \citep{hu2023tifa,cho2024dsg}, GenEval checks object-level properties \citep{ghosh2023geneval}, and VQAScore scores whole-prompt alignment \citep{lin2024vqascore}; and unedited generation already drops subjects \citep{chefer2023attend}. These studies motivate the separate measurements of subject presence and target-description match used in our controlled comparison. Optimising a proxy past the range where it tracks the goal can degrade the goal \citep{manheim2018goodhart,skalse2022hacking,gao2023overopt}, LLMs iterating on feedback hack in-context rewards \citep{pan2024feedback,pan2024selfrefine}, and reward fine-tuning of diffusion models over-optimises \citep{black2024ddpo,clark2024draft,zhang2024overopt}; our rewrite results show a corresponding divergence between the selection score and the task check at the level of weight programs.

\section{Discussion and Conclusion}\label{sec:7}

\textbf{Construction and score-guided revision behave differently.} In this comparison, Astra's one-shot programs achieved higher fresh-scene joint success than the shared-panel baseline, and their code shares one structural pattern: hard projection onto the complement of a protected text-state subspace. The rewrite stage often referenced these direction sets. Yet the resulting Astra updates lost the measured gain while their CLIP scores increased. The hand-written extension under CLIP-guided fitting likewise raised CLIP without a detectable fresh-scene joint-success gain. Access to useful program components was therefore not enough for a CLIP-guided rewrite-and-selection pipeline to keep their task benefit.

The programs' shared preservation structure motivates a component-level question. Our implemented transplant into the hand-written extension projected both the regression inputs and the target residuals. Matrix checks on the two fixed test edits confirm that this two-sided variant differs materially from the registered operator. It changed fresh-scene joint success by +1.0 [-3.6, +5.6] and moved native joint generality by -7.2 points; it did not test the registered input-only transplant, and the complete program's gain remains unattributed to an individual component (Appendix~\ref{app:E}). Two design questions follow: how to keep useful constraints when directions are modified and combined, and how to select updates using direct checks of subject presence and target-description match rather than the CLIP similarity the fitter already used.

\textbf{Scope.} We study Stable Diffusion 1.4, TIMED, one primary VLM judge (with five further evaluators on a subsample) and one program per edit and model, with a single rewrite round. We do not compare against published null-space editors such as SPEED or ACE. The human labels address subject presence and the local attribute question; we report their mixed method-level readings in Appendix~\ref{app:F}.

\textbf{Conclusion.} LLM-written programs can construct useful diffusion updates under a shared fitting protocol. In this study, Astra's one-shot programs improved judged fresh-scene target fulfilment, while score-guided rewriting increased CLIP without retaining that gain. The evaluator panel and local checks define the scope of this observation. Understanding WSP therefore requires examining both what the generated program constructs and what changes after score-guided revision.

\clearpage
\bibliography{references}
\bibliographystyle{wsparxiv}
\clearpage
\appendix
\onecolumn
\raggedbottom

\section*{Impact Statement} This work studies how language models edit the weights of a text-to-image model and how such edits are evaluated. Weight editing can be used to change what a deployed model depicts; our results argue for evaluating such edits with task-level checks of the edited subject rather than a single proxy score. We will release programs, prompts and per-image results, but not edited model weights.

\textbf{LLM use.} LLMs are the object of study: Astra and Opus write programs, Gemini writes fresh prompts, and Qwen is the judge. LLM assistants were also used for code review and drafting. All numbers are rendered from the result files, which will be released.

\section{Statistics and additional analyses}\label{app:A}

\textbf{Resampling.} Contrasts in the main comparison and the conventional-editor follow-up are paired at the edit level. For each edit, we compute a unit's endpoint as the mean over its prompts and 12 seeds and take the difference between two units. We then resample the 24 (main comparison) or 23 (follow-up object stratum) edit-level differences with replacement, B = 50,000 times. The random seed is derived from the contrast's name, so each interval is reproducible on its own. Intervals are percentile intervals, reported to one decimal because endpoints lie on a coarse grid at this n.

Multiplicity in the exploratory matrix (Table~\ref{tab:A6}) uses paired sign-flip randomisation p-values with 200,000 random sign vectors. For the headline contrast, the exact enumeration over all 2$^2$$^4$ sign vectors agrees to the third decimal. The matrix covers every pair among the nine evaluated units on eight endpoints: native joint efficacy and generality, fresh joint success, fresh CLIP generality, fresh-negative specificity, fresh subject presence, and native CLIP efficacy and generality. Holm and Benjamini--Hochberg adjustments are reported within each endpoint (36 contrasts) and over all cells. The follow-up's preregistered family uses paired bootstrap p-values with Holm, as preregistered.

Some analyses use other resampling schemes. The development-panel comparison (Table~\ref{tab:A1}) has n = 16 edits. The prompt-overlap analysis resamples edits with their prompts kept together (20,000 resamples, prompt-weighted means). The judge-selection interval resamples image clusters. Every number in this paper is rendered from the result files by one script, to be released with them, which also lists, for each number, the result key, endpoint, prompt population, contrast and sample.

\begin{figure}[H]
\centering
\includegraphics[width=\linewidth]{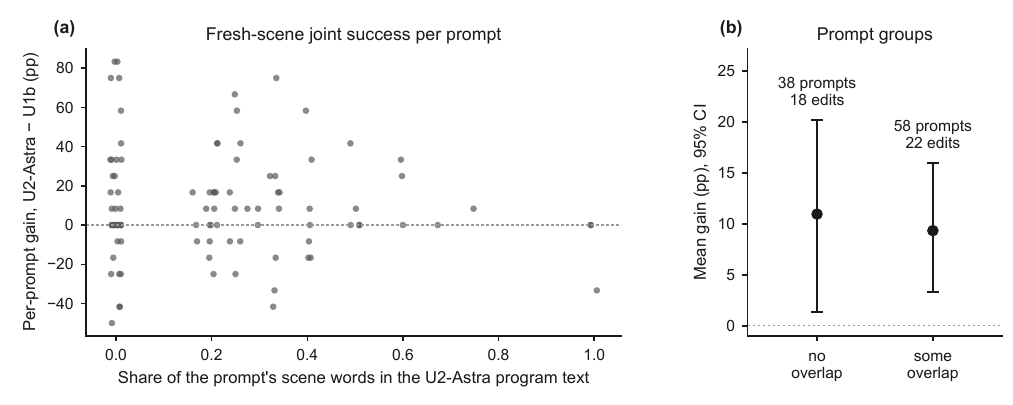}
\caption{Prompt overlap and the construction gain. (a) Per fresh prompt, the U2-Astra $-$ U1b gain in fresh-scene joint success against the share of the prompt's scene words that occur in the U2-Astra program text; horizontal jitter is for visibility. (b) Mean gain on prompts with no and with some literal overlap, with edit-cluster bootstrap intervals; the two groups can share edits.}
\label{fig:A1}
\end{figure}

\textbf{Prompt overlap.} A program writing many scene templates could share scene words with the fresh prompts. On the 38 fresh prompts (18 edits) with no literal scene-word overlap with the matching U2-Astra program the gain is +11.0 [+1.4, +20.2], against +9.3 [+3.3, +16.0] where there is overlap; Opus programs overlap more (0.26 vs 0.22) and show no clear gain (Fig.~\ref{fig:A1}). Lexical overlap is therefore an unlikely sole explanation, though semantic similarity and richer contexts remain.

\textbf{Unit means.} Fig.~\ref{fig:A3} places every unit of the main comparison by its mean fresh-scene CLIP relative match and joint success, including the conventional controls and the target-prompt reference.

\begin{figure}[H]
\centering
\includegraphics[width=4.06in]{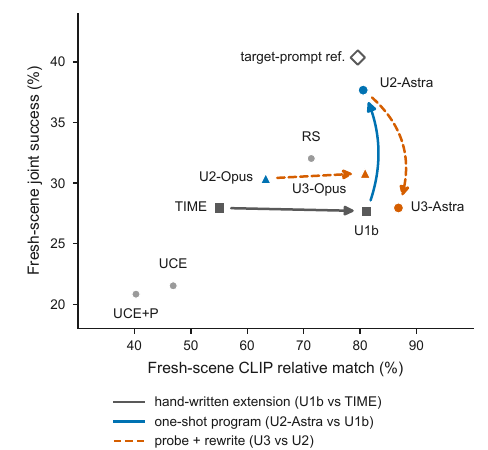}
\caption{Program construction and score-guided rewriting on fresh scenes. Points are mean scores over the 24 TIMED test edits. The blue solid arrow compares U2-Astra with U1b; orange dashed arrows compare U3 with U2 within each model; the grey arrow compares the hand-written extension U1b with TIME. Marker shape marks the model (circle Astra, triangle Opus). Joint success requires subject presence and whole-image target-description matching; the target-prompt reference is the unedited model drawing the target prompt. CLIP increases after rewriting for both models; joint success decreases for Astra and remains inconclusive for Opus. Paired differences and their 95\% edit-bootstrap intervals are in Table~\ref{tab:1}. Fig.~\ref{fig:1} shows the same contrasts as paired changes.}
\label{fig:A3}
\end{figure}

\begin{table*}[h]
\centering\footnotesize\setlength{\tabcolsep}{4pt}
\caption{U2-Astra $-$ U1b on fresh-scene positives by subject type and judge question (top) and by judge threshold (bottom, all 24 edits, VLM joint success; ``soft'' = mean of $\min(p_\mathrm{subject}, p_\mathrm{target})$). Object: the 18 object edits; other: 6 person/brand edits.}
\label{tab:2}
\begin{tabular}{lccc}
\toprule
endpoint & all (24) & object (18) & other (6) \\
\midrule
VLM joint success & \textbf{+10.0\,[+4.7, +15.4]} & \textbf{+7.2\,[+1.7, +12.7]} & \textbf{+18.4\,[+7.3, +29.5]} \\
VLM subject present & +4.8\,[$-$4.8, +14.4] & +0.2\,[$-$9.6, +9.0] & +18.4\,[$-$2.4, +42.4] \\
VLM matches target description & \textbf{+8.9\,[+4.2, +14.1]} & \textbf{+6.7\,[+2.1, +11.3]} & \textbf{+15.6\,[+3.5, +28.5]} \\
CLIP relative match & $-$0.6\,[$-$7.3, +6.4] & $-$1.0\,[$-$6.8, +4.1] & +0.7\,[$-$19.4, +24.7] \\
CLIP success $\wedge$ subject & +3.4\,[$-$4.4, +11.1] & $-$1.0\,[$-$9.1, +6.8] & \textbf{+16.7\,[+1.4, +31.9]} \\
\bottomrule
\end{tabular}

\vspace{4pt}
\begin{tabular}{lcccccccc}
\toprule
threshold & 0.3 & 0.4 & 0.5 & 0.6 & 0.7 & 0.8 & 0.9 & soft \\
\midrule
gain & +9.2 & +8.2 & +10.0 & +8.9 & +8.5 & +7.4 & +5.6 & +7.7 \\
95\% CI & [+3.3, +15.5] & [+2.5, +14.2] & [+4.7, +15.4] & [+4.0, +13.7] & [+3.6, +13.5] & [+3.2, +11.5] & [+1.6, +9.3] & [+3.0, +12.4] \\
\bottomrule
\end{tabular}
\end{table*}

\textbf{Development panels.} On our 16 self-built development edits, U3-Opus led TIME by +25.5 [+11.5, +40.6] in native joint efficacy. On the pre-drawn TIMED edits, the lead is +3.1 [-7.3, +13.2], and TIME rises from 41.1 to 71.2 (Table~\ref{tab:A1}).

\begin{table}[H]
\centering\footnotesize\setlength{\tabcolsep}{3pt}
\caption{Development panel vs TIMED (native joint efficacy, VLM, \%) Both rows use the same judge (Qwen3.8-27B, v2 question) and the same paired edit-level bootstrap; the two panels contain different edits, so the rows are not paired with each other.}
\label{tab:A1}
\begin{tabular}{p{0.30\linewidth}p{0.12\linewidth}p{0.12\linewidth}p{0.12\linewidth}p{0.24\linewidth}}
\toprule
 & TIME & U1b & U3-Opus & U3-Opus $-$ TIME (paired, 95\% CI) \\
\midrule
development panel (16 self-built edits) & 41.1 & 66.1 & 66.7 & +25.5 [+11.5, +40.6] \\
TIMED test (24 pre-drawn edits, main comparison) & 71.2 & 74.7 & 74.3 & +3.1 [-7.3, +13.2] \\
\bottomrule
\end{tabular}
\end{table}

\section{The program-writing prompts and program codes}\label{app:B}

Both templates are reproduced verbatim from the code that sent them. Fields in braces are filled per edit: \texttt{\{old\}}, \texttt{\{new\}} are the edit pair; \texttt{\{layers\}} lists the 32 editable matrices with their shapes and layer groups; in the rewrite template \texttt{\{subject\}}, \texttt{\{subj\_ref\}}, \texttt{\{drop\_T\}}, \texttt{\{tnorm\}}, \texttt{\{panel\}} carry the per-edit subject string, the two fitter tolerances, the TIME update norm and the protection panel, and \texttt{\{schemes\}}, \texttt{\{singletons\}}, \texttt{\{prev\_code\}}, \texttt{\{prev\_notes\}} carry the probing record and the model's own previous program.

\begin{figure}[H]
\centering
\makeatletter
\renewcommand{\fnum@figure}{Box~1}
\makeatother
\includegraphics[width=6.50in]{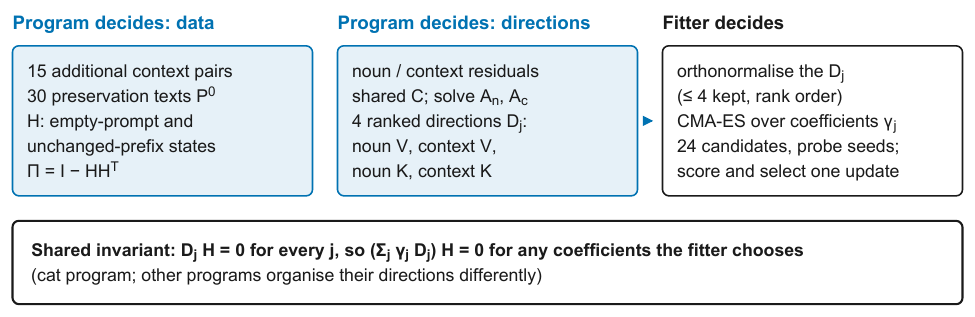}
\caption{Program--fitter responsibilities in the Astra cat example. The program builds a protected subspace $H$ and four ranked directions that all satisfy $D_jH = 0$; the fitter only chooses coefficients, so the invariant holds for every update it can select.}
\addtocounter{figure}{-1}
\label{box:1}
\end{figure}

\paragraph{Verbatim B.1: the one-shot program-writing prompt.}
{\scriptsize\begin{Verbatim}[frame=single,breaklines=true,breakanywhere=true]
You will write a Python program that edits the weights of Stable Diffusion v1.4 (a text-to-image model).

## The edit request
old prompt: "{old}"
new prompt: "{new}"
Goal: after the edit, generating from "{old}" (and from other prompts that mention the same subject, e.g. in
different scenes or styles) should produce images that match "{new}". Prompts about OTHER subjects must keep
producing what the unedited model produces. You do not get any evaluation prompts or scores.

## Where you may edit
The 16 cross-attention layers of the UNet. Each has a key projection W_K and a value projection W_V that map the
CLIP text-encoder token states (77 tokens x 768 dims, last hidden layer) to keys/values. These 32 matrices are
named "v0".."v15" (to_v) and "k0".."k15" (to_k). Layer list (name, module, shape, group):
{layers}

## Tools available inside your program (argument `ctx`, read-only)
- ctx.old, ctx.new: the two prompts.
- ctx.layers: list of dicts with keys name, module, proj, shape, group.
- ctx.W0[name]: the original weight (torch.float32, CPU, shape [out, 768]).
- ctx.encode(list_of_texts) -> torch.float32 tensor [n, 77, 768]: text-encoder states for ANY prompts you write
  (max 64 per call). Positions beyond the end-of-text token hold padding states.
- ctx.tokens(text) -> list of token strings (with <|startoftext|> and <|endoftext|>).
- ctx.align(old_text, new_text) -> list of 77 indices: for each position of old_text, the matching position in
  new_text (the token alignment used by the TIME method).
Allowed imports: torch, numpy, math, re, itertools, functools, collections. No files, network or other modules.
Your program must finish within 60 seconds on CPU.

## Reference (a baseline you should try to beat, not the required answer)
TIME (Orgad et al. 2023) solves, for each matrix W, min_W' sum_t ||W' k_t - W v*_t||^2 + lambda ||W' - W||^2 where
k_t are the old prompt's token states and v*_t the aligned new prompt's token states (summed over the prompt and
3 augmented variants "A photo of / An image of / A picture of ..."), giving
W' = (lambda W + sum_t W v*_t k_t^T)(lambda I + sum_t k_t k_t^T)^(-1), lambda = 0.1. Its known weakness: other
subjects change too (specificity about 68-76%), and generalisation to related prompts is incomplete (about 64-69%).

## What happens with your output
Your directions are ranked. The first 4 are normalised (Frobenius norm over all 32 matrices) and orthogonalised in
your order. A numerical optimiser (CMA-ES) then chooses coefficients beta for these 4 directions, so the final edit is
dW = sum_j beta_j * direction_j, scoring candidates by CLIP similarity of images generated from "{old}" to "{new}"
versus "{old}", minus a penalty for changing images of a fixed set of protection prompts. Directions only need the
right shape and sign/pattern; their scale does not matter. You may also suggest up to 4 protection prompts (prompts
about other subjects whose images should not change) and initial coefficients for your first 4 directions AS YOU
RETURNED THEM (the edit sum_j c_j * direction_j, before normalisation); it is projected onto the basis as a starting
point for the optimiser.

## Output
Reply with your reasoning briefly, then ONE Python code block defining:

def build_directions(ctx):
    ...
    return {{"directions": [dict(name -> torch tensor of that layer's shape), ...],   # ranked, up to 8
             "protect_prompts": [str, ...],                                         # up to 4
             "init_coeffs": [float, ...] or None,                                   # optional, for the first 4
             "notes": "one-paragraph description of the construction"}}
\end{Verbatim}
}

\paragraph{Verbatim B.2: the probe-and-rewrite prompt, appended to B.1 for U3.}
{\scriptsize\begin{Verbatim}[frame=single,breaklines=true,breakanywhere=true]
## Experiment record from forward probing (probe seeds only; no evaluation data)
You previously wrote the program below for this edit. We applied it (directions fitted by the optimiser), and also
measured the TIME baseline, a hand-written conventional improvement "U1b" (TIME + preservation term on the protection
panel + more augmentation templates) and the one-shot programs of two LLMs (U2-astra, U2-opus); yours is {own}.
Images: attached labelled contact sheets (unedited model; each scheme on the old prompt, augmented variants and the
protection panel; each library direction alone at three doses).
Metrics (CLIP ViT-L/14): eff = gain in [sim(image,"{new}") - sim(image,"{old}")] over the unedited model (higher = edit
applied); subj = sim(image, "a photo of {subject}") (the optimiser only accepts subj >= subj_ref - 0.01, subj_ref =
{subj_ref:.4f} being the value for images of "{new}" from the unedited model; lower means the subject is being replaced);
drop = damage on the protection panel (the optimiser only accepts drop <= TIME's drop = {drop_T:.4f} + 0.01). The
protection panel is fixed for this experiment (P1..P{n_panel} in the sheets: {panel}); protect_prompts you return are recorded but not used.
Scale: ||dW_TIME|| = {tnorm:.2f} (Frobenius, all 32 matrices), also available as ctx.time_norm. Dose d of a library
direction = coefficient d * {tnorm:.2f} on the unit-norm ctx.library[name]; init_coeffs are in the same absolute units
(coefficient of each returned direction as you return it). The optimiser searches coefficients of (the orthonormalised
span of) your first 4 independent directions within a Frobenius ball of radius 3 * ||dW_TIME||.

### Whole schemes (as fitted)
{schemes}

### Library directions applied alone (dose = multiple of ||dW_TIME||)
{singletons}

These {n_lib} library directions are available inside your program as ctx.library[name] (dict layer -> tensor, unit Frobenius
norm, holding ALL 32 layer names; layers a direction does not touch are zero tensors): {lib_names}. You may reuse, combine, or replace them; writing entirely new directions is allowed.

### Your previous program
```python
{prev_code}
```
Your previous notes: {prev_notes}

## Task
Rewrite build_directions(ctx) to do better than all schemes above: apply the edit to "{old}" and related prompts
(eff high), keep the subject (subj >= subj_ref) and do not damage other subjects (low drop). Same output format and rules.
\end{Verbatim}
}

\begin{table}[H]
\centering\footnotesize\setlength{\tabcolsep}{3pt}
\caption{Program codes (96 final programs; syntax-tree and pattern codes, \texttt{tools/program\_analysis.py}) Codes are mechanical: function calls and literals from the syntax tree; the projection and library-key codes use patterns over the source. A reference to a library direction in the code does not show that the direction was returned among the four fitted directions or received a non-zero coefficient.}
\label{tab:A5}
\begin{tabular}{p{0.40\linewidth}p{0.12\linewidth}p{0.12\linewidth}p{0.12\linewidth}p{0.12\linewidth}}
\toprule
code & U2-Astra & U2-Opus & U3-Astra & U3-Opus \\
\midrule
calls TIME token alignment (ctx.align) & 24/24 & 24/24 & 3/24 & 16/24 \\
encodes the empty prompt as a text & 24/24 & 0/24 & 20/24 & 0/24 \\
calls an SVD & 24/24 & 0/24 & 22/24 & 0/24 \\
explicit orthogonal projection X $-$ H(H$^\top$X) & 24/24 & 0/24 & 18/24 & 0/24 \\
solves a linear system & 24/24 & 24/24 & 16/24 & 21/24 \\
fitter selected a zero update & 0/24 & 4/24 & 0/24 & 1/24 \\
texts encoded on re-execution (median) & 96 & 198 & 57 & 180 \\
rewrites referencing U2-Astra library directions & --- & --- & 21/24 & 19/24 \\
share of final update outside the probing-library span (median) & --- & --- & 0.51 & 0.26 \\
\bottomrule
\end{tabular}
\end{table}

\section{Fresh-scene prompts}\label{app:C}

Fresh prompts were written after every unit's update was frozen. Before that, every program was re-executed to record all the text it encodes. Gemini-4-argon then wrote, per edit, four positives that put the subject in a new scene or style, each with its edited version, and eight near-neighbour negatives: other subjects that are similar to the subject or often shown with it, with the same change applied. It was told to avoid templates such as "a photo of". Candidates that matched any recorded program text, TIME/U1b template or panel prompt were discarded. Opus then checked every pair. A positive must depict the subject, and its edited version must change only the attribute. A negative must not be the subject or a kind of it, and its edited version must apply the change to that negative. Rejected candidates were regenerated until each edit had 4 + 8 accepted prompts. The accepted prompts were written once and will be released.

\section{Judge questions}\label{app:D}

\begin{itemize}
\item \textbf{Subject, objects:} "Is there $\langle$subject$\rangle$ in this image? Its colour, material, surface pattern or accessories may differ from usual, and it may be drawn in any style, but it must still be $\langle$subject$\rangle$, not a different kind of object."
\item \textbf{Subject, people and characters:} "Is $\langle$subject$\rangle$ (the actual person or character, not someone else) in this image? Hair colour, clothing or accessories may differ, and the depiction may be stylised."
\item \textbf{Target:} "Does this image match the description $\langle$target prompt$\rangle$?" This is a check on the whole image, so joint success means \emph{subject present and the image as a whole matches the target description}; it does not verify that the attribute is bound to the subject.
\end{itemize}

The judge is Qwen3.8-27B, and a "yes" probability of at least 0.5 counts as yes. It was selected on a separate set of 274 generic multi-predicate questions (Table~\ref{tab:A2}). The subject question above replaced a first wording after a 48-image check found that wording too strict on recoloured subjects. Neither question above was re-run on the 274 selection questions.

\begin{table}[H]
\centering\footnotesize\setlength{\tabcolsep}{3pt}
\caption{Judge selection (274 human-labelled questions, 267 with definite human answers used for scoring over 6 predicate families, 189 images, 4 dev edits) Pre-set acceptance rule: macro balanced accuracy $\geq$ 0.90 and every predicate family $\geq$ 0.85. The user saw neither candidate's answers while labelling. These are generic multi-predicate questions (e.g. whether a named subject is in the image). The two questions used in the development experiments and the main comparison --- the reworded subject question and the target-description question (Appendix~\ref{app:D}) --- were not re-run on these 274 questions; the reworded subject question was checked on 48 images only.}
\label{tab:A2}
\begin{tabular}{p{0.34\linewidth}p{0.26\linewidth}p{0.15\linewidth}p{0.15\linewidth}}
\toprule
candidate & macro balanced accuracy [95\% CI] & accuracy & abstain rate \\
\midrule
Qwen3.8-27B, threshold 0.5 (selected) & 0.955 [0.904, 0.977] & 0.914 & 0.000 \\
Qwen3.8-27B, abstain band 0.3--0.7 & 0.900 [0.816, 0.947] & 0.865 & 0.097 \\
Gemini-4-argon & 0.864 [0.784, 0.940] & 0.850 & 0.082 \\
\bottomrule
\end{tabular}
\end{table}

\section{Transplanting the hard constraint}\label{app:E}

The transplant replaces the shared map of the hand-written extension, $M = (E-K)B^{-1}$, with a constrained solve that removes a protected text-state subspace $H$, built the way the Astra programs build it (the empty prompt's states and the unchanged causal prefixes of the training pairs). Here $E$, $K$ and $P_\Sigma$ are the sums of target--source cross-covariances, source covariances and preservation-state covariances, $B = \lambda I + K + \alpha P_\Sigma$ and $B_\Pi = \lambda I + \Pi(K + \alpha P_\Sigma)\Pi$. The registered operator projects the regression inputs and the solution, $M_{\rm reg} = (E-K)\Pi B_\Pi^{-1}\Pi$ with $\Pi = I - HH^\top$. The implementation also projected the target residuals, which yields $M_{\rm impl} = \Pi M_{\rm reg}$: an additional left projection that removes the residual's component in $H$. Both operators annihilate $H$ and both reduce to the unconstrained map when $H$ is empty, which is why the run's validity checks did not separate them; the final right projection is redundant in exact arithmetic for both constrained solves. The reported run is therefore the stricter variant. On the primary judge it changed fresh-scene joint success by +1.0 [-3.6, +5.6] (sign-flip p = 0.72) and native-prompt joint generality by -7.2 [-11.0, -3.3] (p = 0.002). On the bed and cat edits used in the original validity checks, fixed before any difference was seen, the relative Frobenius differences between the two sets of weight updates were 73 and 71\%, measured as the norm of their difference divided by the registered update's norm across all editable matrices. In the shared-map space, the extra output-side projection removed 47 and 45\% of the squared Frobenius norm of $M_{\rm reg}$. The implemented maps had norms 0.76 and 0.78 times those of the unconstrained maps, compared with 1.04 and 1.05 for the registered maps. These matrix checks establish a material change in the update construction. The image-level results reported here belong to the two-sided variant; the input-only transplant remains unevaluated.

{\sloppy\paragraph{Development-stage protection-component study.} This study transplanted structural protection components into a conventional closed-form editor on the 16 development edits. The preregistered curve statistic was found to be biased (it is non-zero when a curve is compared with itself), so the preregistered criterion is reported as not met. A post-hoc same-$\alpha$ comparison suggests a protection gain of about 3 points for one variant with efficacy not distinguishable; it is not used for any claim. Its prefix component was a soft penalty, not the hard projection the Astra programs use, so it does not test that construction.\par}

\section{Conventional-editor follow-up: tests and human labels}\label{app:F}

Raters saw each image with the subject name and the attribute word only, never the target sentence, so their second question corresponds to \texttt{local} and not to the main endpoint. Two raters labelled the same stratified sample from the conventional-editor follow-up; the first did so before the rubric was frozen. Agreement on questions both answered definitely is $\kappa$ = 0.647 (110 images) for subject presence and 0.735 (113) for the attribute, 0.600 and 0.711 when weighted by inclusion probability. Among CLIP-successful images with definite human subject judgements, the inverse-inclusion-probability-weighted absence rates are 19.86\% and 18.25\%.

One rater also labelled the 144-image subsample of Table~\ref{tab:3}. On it, with unsure counted as unsuccessful, the human \texttt{local} reading gives +4.2 [-10.4, +18.8] for U2-Astra $-$ U1b and +0.0 [-18.8, +18.8] for U3-Astra $-$ U2-Astra, against +20.8 [+4.2, +37.5] and -20.8 [-37.5, -4.2] from the primary judge's \texttt{full} reading on the same images. These are different target questions, and we report them for completeness rather than as a test of the judge. Holding the primary judge's target answer fixed and substituting the rater's subject answer where it is definite gives +16.7 and -12.5 (+12.5 and -10.4 if the rater's unsure answers count as absent): the subject-question disagreement accounts for part of this sample's gap and does not remove either sign. This mixes two instruments and is a sensitivity analysis, not a corrected estimate. On subject presence alone, where the two questions coincide in intent but not in wording, the judge calls the subject absent where the rater sees it on 12, 9 and 13 of 44 images for U1b, U2-Astra and U3-Astra.

\begin{table}[H]
\centering\footnotesize\setlength{\tabcolsep}{3pt}
\caption{Preregistered follow-up contrasts (object stratum, n = 23 edits; Holm, $\alpha$ = 0.05) p-values are two-sided paired-bootstrap values (B = 50000) floored at 1/(B+1).}
\label{tab:A3}
\begin{tabular}{p{0.35\linewidth}p{0.25\linewidth}p{0.11\linewidth}p{0.11\linewidth}p{0.11\linewidth}}
\toprule
contrast & mean [95\% CI] (pp) & p & Holm p & rejected \\
\midrule
CLIP generality advantage, rule editor $-$ TIME & +21.4 [+14.9, +28.6] & 2.0e-05 & 1.4e-04 & yes \\
subject presence, rule editor $-$ TIME & -14.5 [-24.2, -5.6] & 3.8e-04 & 1.5e-03 & yes \\
attenuation, rule editor $-$ TIME & +21.7 [+13.5, +30.6] & 2.0e-05 & 1.4e-04 & yes \\
attenuation, TIME $-$ target-prompt ref. & -10.4 [-16.8, -4.5] & 1.0e-04 & 5.0e-04 & yes \\
attenuation, rule editor $-$ target-prompt ref. & +11.3 [+4.0, +19.1] & 1.6e-03 & 4.9e-03 & yes \\
subject-filtered CLIP success, TIME $-$ target-prompt ref. (both filtered) & -8.6 [-15.1, -1.5] & 1.9e-02 & 1.9e-02 & yes \\
subject-filtered CLIP success, rule editor $-$ target-prompt ref. (both filtered) & -9.0 [-16.8, -2.2] & 7.7e-03 & 1.5e-02 & yes \\
\bottomrule
\end{tabular}
\end{table}
\begin{figure}[H]\centering\includegraphics[width=\linewidth]{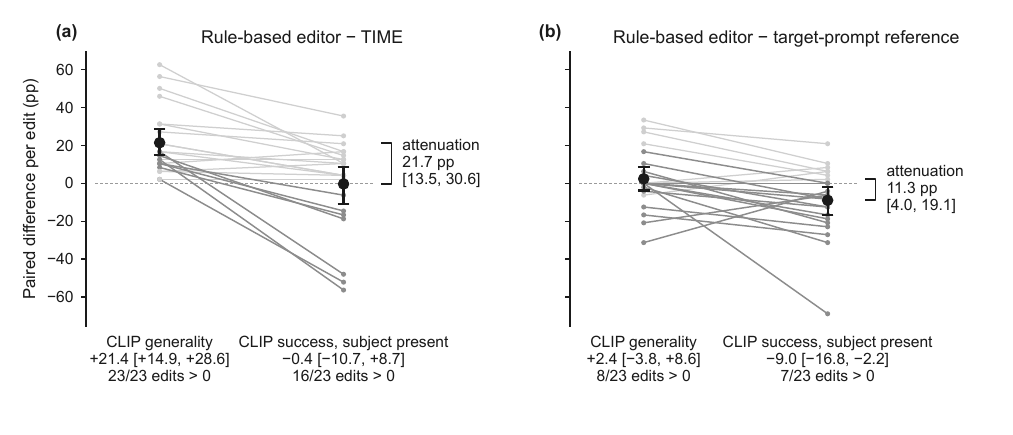}\caption{Conventional-editor follow-up on 23 object edits: the rule-based editor against TIME (a) and against the target-prompt reference (b). Each line is one edit, darker where the subject-filtered difference is zero or below; black: mean and 95\% paired bootstrap interval, printed under each column.}\label{fig:A2}\end{figure}
\begin{table}[H]
\centering\footnotesize\setlength{\tabcolsep}{3pt}
\caption{Human annotations for the conventional-editor follow-up (first of two raters, the second in Appendix~\ref{app:F}; 128 images stratified by arm $\times$ judge verdict from the follow-up's object-stratum fresh-scene positives; reference, not ground truth) Rates are inclusion-probability weighted to the follow-up's object-stratum fresh-positive images. 'Unsure' answers are dropped from the denominator of that question only, so the columns have different denominators and a joint rate can exceed a marginal one. The judge column is the judge on the same weighted population.}
\label{tab:A4}
\begin{tabular}{p{0.17\linewidth}p{0.15\linewidth}p{0.15\linewidth}p{0.15\linewidth}p{0.15\linewidth}p{0.15\linewidth}}
\toprule
arm & human: subject present & human: attribute on subject & human: both & judge: subject present & judge-absent $\rightarrow$ human-present \\
\midrule
unedited & 98.3 & 18.7 & 16.2 & 83.5 & 9/16 \\
target-prompt ref. & 80.0 & 76.6 & 60.9 & 67.3 & 7/16 \\
TIME & 90.7 & 46.1 & 46.7 & 70.5 & 9/16 \\
rule editor & 64.3 & 55.3 & 40.1 & 58.6 & 4/16 \\
\bottomrule
\end{tabular}
\end{table}

\section{Measurement sensitivity: an earlier second evaluator}\label{app:G}

Before the panel of \S\ref{sec:4}, \texttt{gpt-6.1-sol} re-evaluated a six-seed sample of all seven LLM and conventional units, asking both questions in one call. On that sample the primary judge gives +12.2 [+6.1, +18.8] for U2-Astra $-$ U1b, -9.4 [-16.7, -2.1] for U3-Astra $-$ U2-Astra, +1.6 [-6.2, +8.9] for U3-Opus $-$ U2-Opus and -2.3 [-8.9, +4.2] for U1b $-$ TIME; GPT gives +1.9 [-3.1, +6.4], -1.9 [-9.5, +5.9], +4.9 [-3.8, +14.1] and +8.0 [+2.3, +13.9]. The two evaluations differ in sample, in call structure and in image encoding. On the 78 images (21 edits) they share, GPT's U2-Astra $-$ U1b is +0.0 with both questions in one call and +0.0 with separate calls, and its U3-Astra $-$ U2-Astra is +4.8 and -4.8. Splitting the questions therefore did not restore the construction gain on the shared images; the data do not identify which difference between the two evaluations explains the rest. The point estimates above are recomputed from the per-image answers, which will be released; the intervals are those of the run's statistics file, whose bootstrap seed was not recorded.

\section{Run configuration and artifacts}\label{app:H}

\textbf{Evaluator sample.} For each edit, we sampled two of its four fresh-scene prompts without replacement and one of the twelve seeds for each selected prompt, using an edit-specific deterministic random generator. U1b, U2-Astra and U3-Astra share these slots, giving 144 images. Selection did not condition on CLIP or judge outcomes; the sample was drawn once and kept after the primary-judge scores on it were seen.

\textbf{Run configuration.} Images are generated with Stable Diffusion v1.4 at 50 sampling steps and guidance 7.5; probe seeds are 1000--1003 and evaluation seeds 100--111. The CMA-ES fitter uses population 8, seed 0, initial step 0.3$\|$$\Delta$W\_TIME$\|$ and a search ball of radius 3$\|$$\Delta$W\_TIME$\|$; anchors count towards the 24-candidate budget. U1b is TIME with six extra prompt prefixes, a preservation term ($\lambda$ = 0.1, $\alpha$ = 1.0) and one direction per network group (down, mid, first two up blocks, remaining up blocks), keys and values joined within a group. RS samples $\lambda$ \textasciitilde{} LogUniform[0.01, 1], $\alpha$ \textasciitilde{} LogUniform[0.1, 10] and strength \textasciitilde{} Uniform[0.25, 2], and initially retains each extra template, panel prompt and network group independently with probability 0.5, 0.6 and 0.5, respectively; if no network group is selected, one is sampled uniformly. Both selection tolerances are 0.01. The fitter keeps the first four linearly independent returned directions in rank order; program time limits are given with Table~\ref{tab:A7}.

\textbf{Artifact identifiers.} In the artifacts, which will be released, E08 denotes the 24-edit main comparison and E10 the conventional-editor follow-up; T02 denotes the judge-selection set, E05b/E06 the development experiments and E09 the development-stage protection-component study. The judge was selected before E05b. The main-comparison preregistration was first drafted on 8 November 2026 and finalized as v0.3 on 9 November 2026; it will be released with the code; besides the endpoints in Table~\ref{tab:A6} it lists TIMED specificity, validity checks and the panel-dependence contrast, whose results are in the result files to be released.

\begin{table}[H]
\centering\footnotesize\setlength{\tabcolsep}{3pt}
\caption{Cost of the main comparison Input tokens and stage hours are per-model and per-stage totals from the run summary: astra 807,486 in / 575,707 out (51 calls); opus 898,858 in / 375,509 out (78 calls); gemini-4-argon 5,976 in / 22,360 out (25 calls). U3 additionally needs the probing record (median 912 images per edit). Whole main-comparison run: 19.5 wall-clock hours on 8 GPUs. Fitting times are wall time under 8-way parallel load. The program-writing prompt asks for completion within 60 s; the runner enforced 120 s (one-shot) and 180 s (rewrite), and program time includes construction and output validation.}
\label{tab:A7}
\begin{tabular}{p{0.16\linewidth}p{0.18\linewidth}p{0.12\linewidth}p{0.16\linewidth}p{0.14\linewidth}p{0.14\linewidth}}
\toprule
unit & LLM output tokens (all attempts) & first-try success & median program run time (s) & candidate evaluations / edit & median fitting wall time / edit (s) \\
\midrule
TIME & --- & --- & --- & 0 & --- \\
UCE & --- & --- & --- & 0 & 182 \\
UCE+P & --- & --- & --- & 0 & 181 \\
U1b & --- & --- & --- & 24 & 2740 \\
RS & --- & --- & --- & 24 & 990 \\
U2-Astra & 274k & 24/24 & 54 & 24 & 2087 \\
U2-Opus & 168k & 20/24 & 99 & 24 & 1563 \\
U3-Astra & 302k & 21/24 & 1 & 24 & 822 \\
U3-Opus & 181k & 23/24 & 2 & 24 & 833 \\
\bottomrule
\end{tabular}
\end{table}
\begin{table}[H]
\centering\footnotesize\setlength{\tabcolsep}{3pt}
\caption{TIMED template errors Four brand$\rightarrow$pizza edits in the local TIMED copy carry 'sushi' text in gt3/gt4. The conventional-editor follow-up excludes them; the main comparison contains one (Subway). Dropping its two affected prompts changes each unit's native CLIP generality (\%) by: No preregistered contrast changes sign or crosses zero; native joint generality moves by at most 0.3 points when recomputed without the affected prompts.}
\label{tab:A9}
\begin{tabular}{p{0.17\linewidth}p{0.25\linewidth}p{0.25\linewidth}p{0.25\linewidth}}
\toprule
unit & as reported & affected prompts dropped & change \\
\midrule
target-prompt ref. & 92.6 & 92.6 & 0.0 \\
TIME & 63.9 & 63.8 & -0.1 \\
UCE & 49.2 & 50.4 & +1.2 \\
UCE+P & 40.5 & 41.8 & +1.3 \\
U1b & 90.5 & 90.8 & +0.3 \\
RS & 79.8 & 79.6 & -0.2 \\
U2-Astra & 85.1 & 85.6 & +0.5 \\
U2-Opus & 66.9 & 66.7 & -0.2 \\
U3-Astra & 88.6 & 88.7 & +0.1 \\
U3-Opus & 87.0 & 86.7 & -0.3 \\
\bottomrule
\end{tabular}
\end{table}

\section{Complete comparison matrix}\label{app:I}

Table~\ref{tab:A6} lists every unit pair on all eight endpoints of the main comparison, one block per endpoint.
{\footnotesize\setlength{\tabcolsep}{4pt}\setlength{\LTpre}{6pt}\setlength{\LTpost}{2pt}
\Needspace{12\baselineskip}
\begin{longtable}{p{4.0cm}>{\centering\arraybackslash}p{3.1cm}*{5}{>{\centering\arraybackslash}p{1.55cm}}}
\caption{All 288 unit-pair $\times$ endpoint contrasts in the main comparison, one block per endpoint. Difference and 95\% interval: paired bootstrap over 24 edits. $p$: paired sign-flip randomisation (200,000 draws); values below the printed precision are shown as $<$0.0001 ($p$) and $<$0.001 (adjusted). Holm / BH adjusted within the endpoint (36 contrasts) and over all cells. Bold: global BH $<0.05$. $\star$: unit pair and endpoint in the preregistration for the main comparison (their inference follows the preregistration; listed here for completeness).}\label{tab:A6}\\
\toprule
contrast & difference [95\% CI] & $p$ & Holm (end.) & BH (end.) & Holm (all) & BH (all) \\
\midrule
\multicolumn{7}{l}{\textit{native joint efficacy}} \\
\endfirsthead
\toprule
\multicolumn{7}{l}{\textit{Table~\ref{tab:A6} (continued): native joint efficacy}} \\
contrast & difference [95\% CI] & $p$ & Holm (end.) & BH (end.) & Holm (all) & BH (all) \\
\midrule
\endhead
\midrule\multicolumn{7}{r}{\textit{continued on next page}} \\
\endfoot
\endlastfoot
U3-Opus $-$ U3-Astra & +1.0\,[$-$5.9, +9.7] & 0.8939 & 1.000 & 1.000 & 1.000 & 0.950 \\
U3-Opus $-$ U2-Opus $\star$ & +14.6\,[$-$0.7, +31.9] & 0.1120 & 1.000 & 0.237 & 1.000 & 0.189 \\
U3-Opus $-$ U2-Astra & +0.7\,[$-$7.3, +9.7] & 0.9425 & 1.000 & 1.000 & 1.000 & 0.980 \\
U3-Opus $-$ RS & $-$0.7\,[$-$9.0, +8.0] & 0.9381 & 1.000 & 1.000 & 1.000 & 0.979 \\
U3-Opus $-$ U1b $\star$ & $-$0.3\,[$-$6.2, +5.9] & 1.0000 & 1.000 & 1.000 & 1.000 & 1.000 \\
\textbf{U3-Opus $-$ UCE+P $\star$} & +34.7\,[+18.4, +51.0] & 0.0006 & 0.018 & 0.003 & 0.142 & 0.003 \\
\textbf{U3-Opus $-$ UCE $\star$} & +33.7\,[+18.8, +49.0] & 0.0004 & 0.014 & 0.003 & 0.102 & 0.002 \\
U3-Opus $-$ TIME $\star$ & +3.1\,[$-$7.3, +13.2] & 0.6083 & 1.000 & 0.952 & 1.000 & 0.727 \\
U3-Astra $-$ U2-Opus & +13.5\,[$-$3.5, +31.6] & 0.1650 & 1.000 & 0.297 & 1.000 & 0.251 \\
U3-Astra $-$ U2-Astra & $-$0.3\,[$-$9.4, +8.0] & 1.0000 & 1.000 & 1.000 & 1.000 & 1.000 \\
U3-Astra $-$ RS & $-$1.7\,[$-$11.1, +7.3] & 0.7768 & 1.000 & 1.000 & 1.000 & 0.868 \\
U3-Astra $-$ U1b & $-$1.4\,[$-$9.4, +5.9] & 0.8127 & 1.000 & 1.000 & 1.000 & 0.893 \\
\textbf{U3-Astra $-$ UCE+P} & +33.7\,[+15.3, +51.4] & 0.0023 & 0.059 & 0.007 & 0.490 & 0.008 \\
\textbf{U3-Astra $-$ UCE} & +32.6\,[+14.9, +49.7] & 0.0020 & 0.051 & 0.006 & 0.421 & 0.008 \\
U3-Astra $-$ TIME & +2.1\,[$-$9.0, +13.2] & 0.7675 & 1.000 & 1.000 & 1.000 & 0.863 \\
U2-Opus $-$ U2-Astra & $-$13.9\,[$-$30.9, +2.1] & 0.1346 & 1.000 & 0.269 & 1.000 & 0.222 \\
U2-Opus $-$ RS & $-$15.3\,[$-$28.5, $-$3.8] & 0.0263 & 0.631 & 0.073 & 1.000 & 0.062 \\
U2-Opus $-$ U1b & $-$14.9\,[$-$30.9, $-$0.7] & 0.0824 & 1.000 & 0.186 & 1.000 & 0.144 \\
U2-Opus $-$ UCE+P & +20.1\,[+1.4, +38.5] & 0.0515 & 1.000 & 0.133 & 1.000 & 0.100 \\
U2-Opus $-$ UCE & +19.1\,[+0.7, +36.5] & 0.0600 & 1.000 & 0.144 & 1.000 & 0.114 \\
U2-Opus $-$ TIME & $-$11.5\,[$-$26.0, +1.7] & 0.1449 & 1.000 & 0.274 & 1.000 & 0.234 \\
U2-Astra $-$ RS & $-$1.4\,[$-$11.1, +7.6] & 0.8397 & 1.000 & 1.000 & 1.000 & 0.909 \\
U2-Astra $-$ U1b & $-$1.0\,[$-$10.4, +7.3] & 0.8902 & 1.000 & 1.000 & 1.000 & 0.950 \\
\textbf{U2-Astra $-$ UCE+P} & +34.0\,[+18.1, +50.0] & 0.0008 & 0.022 & 0.003 & 0.171 & 0.004 \\
\textbf{U2-Astra $-$ UCE} & +33.0\,[+16.3, +49.0] & 0.0012 & 0.032 & 0.004 & 0.265 & 0.005 \\
U2-Astra $-$ TIME & +2.4\,[$-$8.7, +12.8] & 0.7171 & 1.000 & 1.000 & 1.000 & 0.816 \\
RS $-$ U1b & +0.3\,[$-$7.3, +7.6] & 1.0000 & 1.000 & 1.000 & 1.000 & 1.000 \\
\textbf{RS $-$ UCE+P} & +35.4\,[+21.2, +49.7] & 0.0001 & 0.004 & 0.001 & 0.034 & 0.001 \\
\textbf{RS $-$ UCE} & +34.4\,[+21.2, +47.9] & 0.0001 & 0.003 & 0.001 & 0.019 & 0.001 \\
RS $-$ TIME & +3.8\,[$-$2.1, +10.1] & 0.2801 & 1.000 & 0.480 & 1.000 & 0.384 \\
\textbf{U1b $-$ UCE+P} & +35.1\,[+18.7, +51.0] & 0.0008 & 0.022 & 0.003 & 0.175 & 0.004 \\
\textbf{U1b $-$ UCE} & +34.0\,[+18.7, +49.3] & 0.0005 & 0.016 & 0.003 & 0.124 & 0.003 \\
U1b $-$ TIME $\star$ & +3.5\,[$-$6.2, +12.8] & 0.5369 & 1.000 & 0.879 & 1.000 & 0.661 \\
UCE+P $-$ UCE & $-$1.0\,[$-$6.9, +4.5] & 0.8259 & 1.000 & 1.000 & 1.000 & 0.903 \\
\textbf{UCE+P $-$ TIME} & $-$31.6\,[$-$42.7, $-$20.1] & $<$0.0001 & 0.002 & 0.001 & 0.012 & $<$0.001 \\
\textbf{UCE $-$ TIME} & $-$30.6\,[$-$41.0, $-$20.1] & $<$0.0001 & 0.001 & 0.001 & 0.007 & $<$0.001 \\
\bottomrule
\end{longtable}
\Needspace{12\baselineskip}
\begin{longtable}{p{4.0cm}>{\centering\arraybackslash}p{3.1cm}*{5}{>{\centering\arraybackslash}p{1.55cm}}}

\toprule
contrast & difference [95\% CI] & $p$ & Holm (end.) & BH (end.) & Holm (all) & BH (all) \\
\midrule
\multicolumn{7}{l}{\textit{Table~\ref{tab:A6}, block 2 of 8: native joint generality}} \\
\endfirsthead
\toprule
\multicolumn{7}{l}{\textit{Table~\ref{tab:A6} (continued): native joint generality}} \\
contrast & difference [95\% CI] & $p$ & Holm (end.) & BH (end.) & Holm (all) & BH (all) \\
\midrule
\endhead
\midrule\multicolumn{7}{r}{\textit{continued on next page}} \\
\endfoot
\endlastfoot
U3-Opus $-$ U3-Astra & $-$2.4\,[$-$6.0, +1.0] & 0.2180 & 1.000 & 0.280 & 1.000 & 0.315 \\
U3-Opus $-$ U2-Opus $\star$ & +3.0\,[$-$6.4, +12.8] & 0.5650 & 1.000 & 0.656 & 1.000 & 0.689 \\
\textbf{U3-Opus $-$ U2-Astra} & $-$7.0\,[$-$12.1, $-$1.7] & 0.0189 & 0.362 & 0.038 & 1.000 & 0.046 \\
U3-Opus $-$ RS & $-$1.2\,[$-$8.4, +6.4] & 0.7803 & 1.000 & 0.802 & 1.000 & 0.868 \\
\textbf{U3-Opus $-$ U1b $\star$} & $-$6.0\,[$-$10.3, $-$1.6] & 0.0181 & 0.362 & 0.038 & 1.000 & 0.046 \\
\textbf{U3-Opus $-$ UCE+P $\star$} & +18.5\,[+7.4, +29.9] & 0.0044 & 0.106 & 0.012 & 0.875 & 0.014 \\
\textbf{U3-Opus $-$ UCE $\star$} & +17.8\,[+7.8, +28.3] & 0.0031 & 0.080 & 0.010 & 0.627 & 0.010 \\
U3-Opus $-$ TIME $\star$ & +4.2\,[$-$4.4, +13.3] & 0.3765 & 1.000 & 0.452 & 1.000 & 0.488 \\
U3-Astra $-$ U2-Opus & +5.4\,[$-$2.2, +13.4] & 0.2035 & 1.000 & 0.272 & 1.000 & 0.300 \\
U3-Astra $-$ U2-Astra & $-$4.6\,[$-$9.9, +0.8] & 0.1178 & 1.000 & 0.193 & 1.000 & 0.197 \\
U3-Astra $-$ RS & +1.2\,[$-$4.7, +7.4] & 0.7162 & 1.000 & 0.758 & 1.000 & 0.816 \\
U3-Astra $-$ U1b & $-$3.5\,[$-$8.4, +1.2] & 0.1774 & 1.000 & 0.256 & 1.000 & 0.268 \\
\textbf{U3-Astra $-$ UCE+P} & +21.0\,[+10.4, +32.0] & 0.0012 & 0.033 & 0.004 & 0.271 & 0.005 \\
\textbf{U3-Astra $-$ UCE} & +20.3\,[+10.7, +30.2] & 0.0006 & 0.017 & 0.003 & 0.137 & 0.003 \\
U3-Astra $-$ TIME & +6.7\,[$-$1.6, +15.3] & 0.1471 & 1.000 & 0.224 & 1.000 & 0.234 \\
U2-Opus $-$ U2-Astra & $-$10.0\,[$-$19.8, $-$0.6] & 0.0607 & 1.000 & 0.109 & 1.000 & 0.114 \\
U2-Opus $-$ RS & $-$4.2\,[$-$12.4, +3.6] & 0.3437 & 1.000 & 0.427 & 1.000 & 0.456 \\
U2-Opus $-$ U1b & $-$9.0\,[$-$18.4, $-$0.1] & 0.0774 & 1.000 & 0.133 & 1.000 & 0.138 \\
\textbf{U2-Opus $-$ UCE+P} & +15.6\,[+4.6, +27.2] & 0.0153 & 0.321 & 0.034 & 1.000 & 0.040 \\
\textbf{U2-Opus $-$ UCE} & +14.9\,[+5.2, +25.2] & 0.0076 & 0.175 & 0.020 & 1.000 & 0.022 \\
U2-Opus $-$ TIME & +1.2\,[$-$7.7, +10.1] & 0.8016 & 1.000 & 0.802 & 1.000 & 0.885 \\
U2-Astra $-$ RS & +5.8\,[+0.6, +11.0] & 0.0454 & 0.817 & 0.086 & 1.000 & 0.091 \\
U2-Astra $-$ U1b & +1.0\,[$-$4.0, +6.1] & 0.7161 & 1.000 & 0.758 & 1.000 & 0.816 \\
\textbf{U2-Astra $-$ UCE+P} & +25.6\,[+16.2, +35.3] & $<$0.0001 & 0.001 & $<$0.001 & 0.006 & $<$0.001 \\
\textbf{U2-Astra $-$ UCE} & +24.9\,[+16.3, +33.9] & $<$0.0001 & 0.001 & $<$0.001 & 0.007 & $<$0.001 \\
\textbf{U2-Astra $-$ TIME} & +11.2\,[+4.6, +17.9] & 0.0042 & 0.105 & 0.012 & 0.845 & 0.014 \\
RS $-$ U1b & $-$4.8\,[$-$11.9, +1.9] & 0.2039 & 1.000 & 0.272 & 1.000 & 0.300 \\
\textbf{RS $-$ UCE+P} & +19.7\,[+10.3, +29.4] & 0.0007 & 0.019 & 0.003 & 0.153 & 0.003 \\
\textbf{RS $-$ UCE} & +19.0\,[+10.6, +27.9] & 0.0003 & 0.008 & 0.001 & 0.062 & 0.002 \\
RS $-$ TIME & +5.4\,[$-$1.5, +12.2] & 0.1491 & 1.000 & 0.224 & 1.000 & 0.236 \\
\textbf{U1b $-$ UCE+P} & +24.5\,[+15.3, +33.4] & 0.0001 & 0.003 & 0.001 & 0.022 & 0.001 \\
\textbf{U1b $-$ UCE} & +23.8\,[+15.8, +31.7] & $<$0.0001 & 0.001 & $<$0.001 & 0.009 & $<$0.001 \\
\textbf{U1b $-$ TIME $\star$} & +10.2\,[+3.3, +17.5] & 0.0110 & 0.242 & 0.026 & 1.000 & 0.031 \\
UCE+P $-$ UCE & $-$0.7\,[$-$3.1, +1.9] & 0.6418 & 1.000 & 0.722 & 1.000 & 0.757 \\
\textbf{UCE+P $-$ TIME} & $-$14.3\,[$-$20.2, $-$8.7] & 0.0001 & 0.003 & 0.001 & 0.024 & 0.001 \\
\textbf{UCE $-$ TIME} & $-$13.6\,[$-$18.5, $-$9.2] & $<$0.0001 & 0.001 & $<$0.001 & 0.006 & $<$0.001 \\
\bottomrule
\end{longtable}
\Needspace{12\baselineskip}
\begin{longtable}{p{4.0cm}>{\centering\arraybackslash}p{3.1cm}*{5}{>{\centering\arraybackslash}p{1.55cm}}}

\toprule
contrast & difference [95\% CI] & $p$ & Holm (end.) & BH (end.) & Holm (all) & BH (all) \\
\midrule
\multicolumn{7}{l}{\textit{Table~\ref{tab:A6}, block 3 of 8: fresh joint success}} \\
\endfirsthead
\toprule
\multicolumn{7}{l}{\textit{Table~\ref{tab:A6} (continued): fresh joint success}} \\
contrast & difference [95\% CI] & $p$ & Holm (end.) & BH (end.) & Holm (all) & BH (all) \\
\midrule
\endhead
\midrule\multicolumn{7}{r}{\textit{continued on next page}} \\
\endfoot
\endlastfoot
U3-Opus $-$ U3-Astra & +2.9\,[$-$0.8, +6.4] & 0.1547 & 1.000 & 0.278 & 1.000 & 0.240 \\
U3-Opus $-$ U2-Opus $\star$ & +0.4\,[$-$6.8, +7.5] & 0.9267 & 1.000 & 0.988 & 1.000 & 0.971 \\
U3-Opus $-$ U2-Astra & $-$6.9\,[$-$12.7, $-$1.5] & 0.0283 & 0.794 & 0.100 & 1.000 & 0.065 \\
U3-Opus $-$ RS & $-$1.2\,[$-$8.2, +4.8] & 0.7578 & 1.000 & 0.853 & 1.000 & 0.856 \\
U3-Opus $-$ U1b $\star$ & +3.1\,[$-$3.0, +8.6] & 0.3316 & 1.000 & 0.477 & 1.000 & 0.442 \\
U3-Opus $-$ UCE+P $\star$ & +10.0\,[+1.7, +18.4] & 0.0332 & 0.854 & 0.100 & 1.000 & 0.071 \\
U3-Opus $-$ UCE $\star$ & +9.3\,[+1.3, +17.4] & 0.0394 & 0.946 & 0.109 & 1.000 & 0.082 \\
U3-Opus $-$ TIME $\star$ & +2.9\,[$-$4.1, +9.8] & 0.4497 & 1.000 & 0.623 & 1.000 & 0.571 \\
U3-Astra $-$ U2-Opus & $-$2.4\,[$-$9.5, +4.5] & 0.5287 & 1.000 & 0.656 & 1.000 & 0.656 \\
\textbf{U3-Astra $-$ U2-Astra} & $-$9.7\,[$-$16.6, $-$2.9] & 0.0139 & 0.418 & 0.072 & 1.000 & 0.036 \\
U3-Astra $-$ RS & $-$4.1\,[$-$10.5, +1.6] & 0.2263 & 1.000 & 0.370 & 1.000 & 0.324 \\
U3-Astra $-$ U1b & +0.3\,[$-$6.4, +6.8] & 0.9603 & 1.000 & 0.988 & 1.000 & 0.986 \\
U3-Astra $-$ UCE+P & +7.1\,[$-$1.5, +16.2] & 0.1424 & 1.000 & 0.270 & 1.000 & 0.232 \\
U3-Astra $-$ UCE & +6.4\,[$-$1.7, +15.0] & 0.1633 & 1.000 & 0.280 & 1.000 & 0.250 \\
U3-Astra $-$ TIME & 0.0\,[$-$7.5, +7.4] & 1.0000 & 1.000 & 1.000 & 1.000 & 1.000 \\
U2-Opus $-$ U2-Astra & $-$7.3\,[$-$15.9, +1.0] & 0.1189 & 1.000 & 0.238 & 1.000 & 0.198 \\
U2-Opus $-$ RS & $-$1.6\,[$-$9.2, +5.7] & 0.6928 & 1.000 & 0.805 & 1.000 & 0.803 \\
U2-Opus $-$ U1b & +2.7\,[$-$4.9, +9.5] & 0.5064 & 1.000 & 0.656 & 1.000 & 0.637 \\
U2-Opus $-$ UCE+P & +9.5\,[+1.3, +17.2] & 0.0328 & 0.854 & 0.100 & 1.000 & 0.071 \\
U2-Opus $-$ UCE & +8.9\,[+0.7, +16.4] & 0.0440 & 1.000 & 0.113 & 1.000 & 0.089 \\
U2-Opus $-$ TIME & +2.4\,[$-$4.7, +9.0] & 0.5237 & 1.000 & 0.656 & 1.000 & 0.653 \\
U2-Astra $-$ RS & +5.6\,[0.0, +11.5] & 0.0788 & 1.000 & 0.167 & 1.000 & 0.139 \\
\textbf{U2-Astra $-$ U1b} & +10.0\,[+4.7, +15.4] & 0.0016 & 0.051 & 0.014 & 0.343 & 0.007 \\
\textbf{U2-Astra $-$ UCE+P} & +16.8\,[+9.5, +24.1] & 0.0003 & 0.011 & 0.007 & 0.072 & 0.002 \\
\textbf{U2-Astra $-$ UCE} & +16.1\,[+8.8, +23.4] & 0.0004 & 0.013 & 0.007 & 0.090 & 0.002 \\
\textbf{U2-Astra $-$ TIME} & +9.7\,[+3.6, +15.7] & 0.0067 & 0.213 & 0.048 & 1.000 & 0.020 \\
RS $-$ U1b & +4.3\,[$-$2.7, +11.5] & 0.2608 & 1.000 & 0.408 & 1.000 & 0.365 \\
\textbf{RS $-$ UCE+P} & +11.2\,[+2.9, +19.7] & 0.0183 & 0.529 & 0.082 & 1.000 & 0.046 \\
RS $-$ UCE & +10.5\,[+2.0, +19.2] & 0.0286 & 0.794 & 0.100 & 1.000 & 0.065 \\
RS $-$ TIME & +4.1\,[$-$2.9, +10.9] & 0.2801 & 1.000 & 0.420 & 1.000 & 0.384 \\
U1b $-$ UCE+P & +6.9\,[+0.1, +13.5] & 0.0654 & 1.000 & 0.157 & 1.000 & 0.121 \\
U1b $-$ UCE & +6.2\,[$-$0.1, +12.5] & 0.0748 & 1.000 & 0.167 & 1.000 & 0.136 \\
U1b $-$ TIME $\star$ & $-$0.3\,[$-$5.7, +5.3] & 0.9518 & 1.000 & 0.988 & 1.000 & 0.982 \\
UCE+P $-$ UCE & $-$0.7\,[$-$3.0, +1.7] & 0.6355 & 1.000 & 0.763 & 1.000 & 0.753 \\
\textbf{UCE+P $-$ TIME} & $-$7.1\,[$-$10.9, $-$3.6] & 0.0013 & 0.044 & 0.014 & 0.284 & 0.005 \\
\textbf{UCE $-$ TIME} & $-$6.4\,[$-$11.2, $-$2.0] & 0.0134 & 0.417 & 0.072 & 1.000 & 0.036 \\
\bottomrule
\end{longtable}
\Needspace{12\baselineskip}
\begin{longtable}{p{4.0cm}>{\centering\arraybackslash}p{3.1cm}*{5}{>{\centering\arraybackslash}p{1.55cm}}}

\toprule
contrast & difference [95\% CI] & $p$ & Holm (end.) & BH (end.) & Holm (all) & BH (all) \\
\midrule
\multicolumn{7}{l}{\textit{Table~\ref{tab:A6}, block 4 of 8: fresh CLIP generality}} \\
\endfirsthead
\toprule
\multicolumn{7}{l}{\textit{Table~\ref{tab:A6} (continued): fresh CLIP generality}} \\
contrast & difference [95\% CI] & $p$ & Holm (end.) & BH (end.) & Holm (all) & BH (all) \\
\midrule
\endhead
\midrule\multicolumn{7}{r}{\textit{continued on next page}} \\
\endfoot
\endlastfoot
U3-Opus $-$ U3-Astra & $-$5.9\,[$-$14.6, +1.6] & 0.1982 & 1.000 & 0.235 & 1.000 & 0.296 \\
\textbf{U3-Opus $-$ U2-Opus} & +17.6\,[+5.4, +31.9] & 0.0123 & 0.221 & 0.023 & 1.000 & 0.033 \\
U3-Opus $-$ U2-Astra & +0.3\,[$-$7.8, +7.9] & 0.9496 & 1.000 & 0.968 & 1.000 & 0.982 \\
U3-Opus $-$ RS & +9.5\,[+1.1, +18.8] & 0.0498 & 0.597 & 0.072 & 1.000 & 0.098 \\
U3-Opus $-$ U1b & $-$0.3\,[$-$8.2, +6.5] & 0.9677 & 1.000 & 0.968 & 1.000 & 0.988 \\
\textbf{U3-Opus $-$ UCE+P} & +40.6\,[+25.0, +54.9] & 0.0001 & 0.004 & $<$0.001 & 0.034 & 0.001 \\
\textbf{U3-Opus $-$ UCE} & +34.0\,[+18.4, +48.4] & 0.0006 & 0.014 & 0.002 & 0.138 & 0.003 \\
\textbf{U3-Opus $-$ TIME} & +25.8\,[+13.5, +37.8] & 0.0003 & 0.008 & 0.001 & 0.076 & 0.002 \\
\textbf{U3-Astra $-$ U2-Opus} & +23.5\,[+9.5, +38.7] & 0.0042 & 0.083 & 0.009 & 0.844 & 0.014 \\
U3-Astra $-$ U2-Astra & +6.2\,[+1.3, +11.6] & 0.0302 & 0.453 & 0.049 & 1.000 & 0.067 \\
\textbf{U3-Astra $-$ RS} & +15.5\,[+7.0, +24.5] & 0.0016 & 0.033 & 0.004 & 0.349 & 0.007 \\
U3-Astra $-$ U1b & +5.6\,[$-$0.6, +13.3] & 0.1453 & 1.000 & 0.180 & 1.000 & 0.234 \\
\textbf{U3-Astra $-$ UCE+P} & +46.5\,[+34.2, +58.6] & $<$0.0001 & $<$0.001 & $<$0.001 & 0.001 & $<$0.001 \\
\textbf{U3-Astra $-$ UCE} & +39.9\,[+27.0, +52.3] & $<$0.0001 & 0.001 & $<$0.001 & 0.008 & $<$0.001 \\
\textbf{U3-Astra $-$ TIME} & +31.7\,[+22.1, +42.3] & $<$0.0001 & $<$0.001 & $<$0.001 & 0.001 & $<$0.001 \\
U2-Opus $-$ U2-Astra & $-$17.3\,[$-$32.9, $-$3.5] & 0.0349 & 0.489 & 0.055 & 1.000 & 0.074 \\
U2-Opus $-$ RS & $-$8.1\,[$-$20.5, +2.5] & 0.2113 & 1.000 & 0.238 & 1.000 & 0.309 \\
U2-Opus $-$ U1b & $-$17.9\,[$-$34.0, $-$3.1] & 0.0365 & 0.489 & 0.055 & 1.000 & 0.076 \\
U2-Opus $-$ UCE+P & +23.0\,[+3.9, +40.5] & 0.0267 & 0.427 & 0.046 & 1.000 & 0.062 \\
U2-Opus $-$ UCE & +16.4\,[$-$1.3, +33.4] & 0.0849 & 0.764 & 0.109 & 1.000 & 0.147 \\
U2-Opus $-$ TIME & +8.2\,[$-$7.6, +22.0] & 0.3104 & 1.000 & 0.339 & 1.000 & 0.422 \\
U2-Astra $-$ RS & +9.2\,[+0.8, +18.1] & 0.0525 & 0.597 & 0.073 & 1.000 & 0.101 \\
U2-Astra $-$ U1b & $-$0.6\,[$-$7.3, +6.4] & 0.8880 & 1.000 & 0.940 & 1.000 & 0.950 \\
\textbf{U2-Astra $-$ UCE+P} & +40.3\,[+29.3, +51.3] & $<$0.0001 & $<$0.001 & $<$0.001 & 0.001 & $<$0.001 \\
\textbf{U2-Astra $-$ UCE} & +33.7\,[+22.0, +45.1] & $<$0.0001 & 0.001 & $<$0.001 & 0.008 & $<$0.001 \\
\textbf{U2-Astra $-$ TIME} & +25.4\,[+16.8, +34.9] & $<$0.0001 & $<$0.001 & $<$0.001 & 0.004 & $<$0.001 \\
RS $-$ U1b & $-$9.8\,[$-$19.9, $-$0.3] & 0.0691 & 0.691 & 0.092 & 1.000 & 0.127 \\
\textbf{RS $-$ UCE+P} & +31.1\,[+16.3, +44.8] & 0.0007 & 0.015 & 0.002 & 0.154 & 0.003 \\
\textbf{RS $-$ UCE} & +24.5\,[+9.5, +38.6] & 0.0043 & 0.083 & 0.009 & 0.866 & 0.014 \\
\textbf{RS $-$ TIME} & +16.2\,[+9.1, +23.0] & 0.0004 & 0.010 & 0.001 & 0.096 & 0.002 \\
\textbf{U1b $-$ UCE+P} & +40.9\,[+29.0, +52.3] & $<$0.0001 & $<$0.001 & $<$0.001 & 0.003 & $<$0.001 \\
\textbf{U1b $-$ UCE} & +34.3\,[+22.0, +45.9] & $<$0.0001 & 0.001 & $<$0.001 & 0.011 & $<$0.001 \\
\textbf{U1b $-$ TIME} & +26.0\,[+17.1, +36.3] & $<$0.0001 & $<$0.001 & $<$0.001 & 0.001 & $<$0.001 \\
\textbf{UCE+P $-$ UCE} & $-$6.6\,[$-$10.0, $-$3.5] & 0.0005 & 0.012 & 0.001 & 0.113 & 0.003 \\
UCE+P $-$ TIME & $-$14.8\,[$-$25.7, $-$2.8] & 0.0220 & 0.374 & 0.040 & 1.000 & 0.053 \\
UCE $-$ TIME & $-$8.2\,[$-$19.5, +4.3] & 0.2026 & 1.000 & 0.235 & 1.000 & 0.300 \\
\bottomrule
\end{longtable}
\Needspace{12\baselineskip}
\begin{longtable}{p{4.0cm}>{\centering\arraybackslash}p{3.1cm}*{5}{>{\centering\arraybackslash}p{1.55cm}}}

\toprule
contrast & difference [95\% CI] & $p$ & Holm (end.) & BH (end.) & Holm (all) & BH (all) \\
\midrule
\multicolumn{7}{l}{\textit{Table~\ref{tab:A6}, block 5 of 8: fresh-negative CLIP specificity}} \\
\endfirsthead
\toprule
\multicolumn{7}{l}{\textit{Table~\ref{tab:A6} (continued): fresh-negative CLIP specificity}} \\
contrast & difference [95\% CI] & $p$ & Holm (end.) & BH (end.) & Holm (all) & BH (all) \\
\midrule
\endhead
\midrule\multicolumn{7}{r}{\textit{continued on next page}} \\
\endfoot
\endlastfoot
U3-Opus $-$ U3-Astra & +2.0\,[$-$1.9, +6.2] & 0.3607 & 1.000 & 0.481 & 1.000 & 0.477 \\
U3-Opus $-$ U2-Opus $\star$ & $-$5.5\,[$-$11.3, 0.0] & 0.0715 & 1.000 & 0.123 & 1.000 & 0.130 \\
U3-Opus $-$ U2-Astra & $-$4.2\,[$-$9.1, +0.3] & 0.1058 & 1.000 & 0.173 & 1.000 & 0.181 \\
U3-Opus $-$ RS & +3.1\,[$-$2.0, +7.9] & 0.2428 & 1.000 & 0.350 & 1.000 & 0.343 \\
U3-Opus $-$ U1b $\star$ & +1.7\,[$-$2.4, +5.8] & 0.4308 & 1.000 & 0.535 & 1.000 & 0.549 \\
U3-Opus $-$ UCE+P $\star$ & $-$6.6\,[$-$12.1, $-$1.6] & 0.0228 & 0.478 & 0.051 & 1.000 & 0.054 \\
U3-Opus $-$ UCE $\star$ & +7.3\,[$-$2.7, +18.6] & 0.2136 & 1.000 & 0.334 & 1.000 & 0.311 \\
\textbf{U3-Opus $-$ TIME $\star$} & +8.6\,[+3.3, +13.8] & 0.0050 & 0.136 & 0.018 & 0.990 & 0.016 \\
U3-Astra $-$ U2-Opus & $-$7.6\,[$-$14.4, $-$1.0] & 0.0424 & 0.721 & 0.076 & 1.000 & 0.087 \\
U3-Astra $-$ U2-Astra & $-$6.2\,[$-$11.5, $-$1.6] & 0.0209 & 0.461 & 0.050 & 1.000 & 0.051 \\
U3-Astra $-$ RS & +1.1\,[$-$5.5, +7.2] & 0.7558 & 1.000 & 0.800 & 1.000 & 0.856 \\
U3-Astra $-$ U1b & $-$0.3\,[$-$5.3, +4.5] & 0.9190 & 1.000 & 0.919 & 1.000 & 0.966 \\
\textbf{U3-Astra $-$ UCE+P} & $-$8.6\,[$-$14.9, $-$3.0] & 0.0102 & 0.245 & 0.028 & 1.000 & 0.029 \\
U3-Astra $-$ UCE & +5.3\,[$-$5.8, +17.4] & 0.4069 & 1.000 & 0.523 & 1.000 & 0.521 \\
U3-Astra $-$ TIME & +6.6\,[+0.9, +12.0] & 0.0351 & 0.632 & 0.066 & 1.000 & 0.074 \\
U2-Opus $-$ U2-Astra & +1.3\,[$-$3.4, +6.1] & 0.6212 & 1.000 & 0.699 & 1.000 & 0.739 \\
\textbf{U2-Opus $-$ RS} & +8.6\,[+3.9, +13.4] & 0.0023 & 0.068 & 0.011 & 0.490 & 0.008 \\
\textbf{U2-Opus $-$ U1b} & +7.2\,[+3.7, +11.1] & 0.0008 & 0.024 & 0.005 & 0.175 & 0.004 \\
U2-Opus $-$ UCE+P & $-$1.1\,[$-$5.4, +2.9] & 0.6486 & 1.000 & 0.708 & 1.000 & 0.759 \\
\textbf{U2-Opus $-$ UCE} & +12.8\,[+3.5, +23.2] & 0.0166 & 0.381 & 0.043 & 1.000 & 0.042 \\
\textbf{U2-Opus $-$ TIME} & +14.1\,[+8.2, +20.0] & 0.0002 & 0.007 & 0.002 & 0.053 & 0.001 \\
\textbf{U2-Astra $-$ RS} & +7.3\,[+2.7, +12.0] & 0.0072 & 0.187 & 0.023 & 1.000 & 0.022 \\
\textbf{U2-Astra $-$ U1b} & +5.9\,[+2.5, +9.3] & 0.0027 & 0.074 & 0.011 & 0.548 & 0.009 \\
U2-Astra $-$ UCE+P & $-$2.4\,[$-$6.2, +1.3] & 0.2395 & 1.000 & 0.350 & 1.000 & 0.340 \\
U2-Astra $-$ UCE & +11.5\,[+2.6, +22.0] & 0.0268 & 0.535 & 0.057 & 1.000 & 0.062 \\
\textbf{U2-Astra $-$ TIME} & +12.8\,[+8.0, +17.9] & $<$0.0001 & 0.001 & 0.001 & 0.008 & $<$0.001 \\
RS $-$ U1b & $-$1.4\,[$-$6.2, +3.2] & 0.5856 & 1.000 & 0.680 & 1.000 & 0.709 \\
\textbf{RS $-$ UCE+P} & $-$9.7\,[$-$15.1, $-$4.3] & 0.0023 & 0.068 & 0.011 & 0.477 & 0.008 \\
RS $-$ UCE & +4.2\,[$-$6.6, +16.5] & 0.5106 & 1.000 & 0.613 & 1.000 & 0.639 \\
RS $-$ TIME & +5.5\,[+1.0, +9.9] & 0.0301 & 0.572 & 0.060 & 1.000 & 0.067 \\
\textbf{U1b $-$ UCE+P} & $-$8.3\,[$-$12.0, $-$4.7] & 0.0002 & 0.008 & 0.002 & 0.059 & 0.002 \\
U1b $-$ UCE & +5.6\,[$-$3.6, +16.6] & 0.3178 & 1.000 & 0.440 & 1.000 & 0.426 \\
\textbf{U1b $-$ TIME $\star$} & +6.9\,[+2.3, +11.8] & 0.0093 & 0.233 & 0.028 & 1.000 & 0.027 \\
\textbf{UCE+P $-$ UCE} & +13.9\,[+7.1, +22.2] & $<$0.0001 & 0.002 & 0.001 & 0.012 & $<$0.001 \\
\textbf{UCE+P $-$ TIME} & +15.2\,[+9.9, +20.5] & $<$0.0001 & 0.001 & 0.001 & 0.011 & $<$0.001 \\
UCE $-$ TIME & +1.3\,[$-$10.2, +10.9] & 0.8274 & 1.000 & 0.851 & 1.000 & 0.903 \\
\bottomrule
\end{longtable}
\Needspace{12\baselineskip}
\begin{longtable}{p{4.0cm}>{\centering\arraybackslash}p{3.1cm}*{5}{>{\centering\arraybackslash}p{1.55cm}}}

\toprule
contrast & difference [95\% CI] & $p$ & Holm (end.) & BH (end.) & Holm (all) & BH (all) \\
\midrule
\multicolumn{7}{l}{\textit{Table~\ref{tab:A6}, block 6 of 8: fresh subject presence}} \\
\endfirsthead
\toprule
\multicolumn{7}{l}{\textit{Table~\ref{tab:A6} (continued): fresh subject presence}} \\
contrast & difference [95\% CI] & $p$ & Holm (end.) & BH (end.) & Holm (all) & BH (all) \\
\midrule
\endhead
\midrule\multicolumn{7}{r}{\textit{continued on next page}} \\
\endfoot
\endlastfoot
U3-Opus $-$ U3-Astra & +1.1\,[$-$3.8, +6.1] & 0.6899 & 1.000 & 0.892 & 1.000 & 0.803 \\
U3-Opus $-$ U2-Opus & $-$8.6\,[$-$16.1, $-$1.6] & 0.0314 & 1.000 & 0.210 & 1.000 & 0.068 \\
U3-Opus $-$ U2-Astra & $-$6.9\,[$-$15.2, +1.2] & 0.1297 & 1.000 & 0.359 & 1.000 & 0.215 \\
U3-Opus $-$ RS & $-$8.1\,[$-$15.6, $-$0.7] & 0.0528 & 1.000 & 0.211 & 1.000 & 0.101 \\
U3-Opus $-$ U1b & $-$2.1\,[$-$10.4, +5.7] & 0.6440 & 1.000 & 0.892 & 1.000 & 0.757 \\
U3-Opus $-$ UCE+P & $-$9.6\,[$-$19.3, +0.3] & 0.0767 & 1.000 & 0.276 & 1.000 & 0.138 \\
U3-Opus $-$ UCE & $-$7.7\,[$-$16.0, +0.6] & 0.0896 & 1.000 & 0.293 & 1.000 & 0.154 \\
\textbf{U3-Opus $-$ TIME} & $-$10.2\,[$-$17.0, $-$4.1] & 0.0037 & 0.130 & 0.067 & 0.758 & 0.013 \\
\textbf{U3-Astra $-$ U2-Opus} & $-$9.7\,[$-$16.8, $-$3.0] & 0.0122 & 0.414 & 0.146 & 1.000 & 0.033 \\
U3-Astra $-$ U2-Astra & $-$8.0\,[$-$17.2, +1.2] & 0.1106 & 1.000 & 0.332 & 1.000 & 0.189 \\
U3-Astra $-$ RS & $-$9.2\,[$-$16.8, $-$2.0] & 0.0231 & 0.762 & 0.208 & 1.000 & 0.054 \\
U3-Astra $-$ U1b & $-$3.2\,[$-$10.3, +3.4] & 0.3940 & 1.000 & 0.659 & 1.000 & 0.509 \\
U3-Astra $-$ UCE+P & $-$10.8\,[$-$19.6, $-$0.5] & 0.0436 & 1.000 & 0.210 & 1.000 & 0.089 \\
U3-Astra $-$ UCE & $-$8.9\,[$-$16.4, $-$0.6] & 0.0467 & 1.000 & 0.210 & 1.000 & 0.093 \\
\textbf{U3-Astra $-$ TIME} & $-$11.4\,[$-$18.3, $-$4.9] & 0.0022 & 0.079 & 0.067 & 0.468 & 0.008 \\
U2-Opus $-$ U2-Astra & +1.7\,[$-$5.9, +9.9] & 0.6939 & 1.000 & 0.892 & 1.000 & 0.803 \\
U2-Opus $-$ RS & +0.5\,[$-$5.8, +7.1] & 0.9002 & 1.000 & 0.936 & 1.000 & 0.953 \\
U2-Opus $-$ U1b & +6.5\,[$-$1.6, +15.1] & 0.1571 & 1.000 & 0.377 & 1.000 & 0.242 \\
U2-Opus $-$ UCE+P & $-$1.0\,[$-$8.8, +7.6] & 0.8314 & 1.000 & 0.936 & 1.000 & 0.904 \\
U2-Opus $-$ UCE & +0.9\,[$-$6.3, +8.7] & 0.8456 & 1.000 & 0.936 & 1.000 & 0.912 \\
U2-Opus $-$ TIME & $-$1.6\,[$-$6.9, +3.6] & 0.5728 & 1.000 & 0.859 & 1.000 & 0.696 \\
U2-Astra $-$ RS & $-$1.2\,[$-$8.7, +6.4] & 0.7797 & 1.000 & 0.936 & 1.000 & 0.868 \\
U2-Astra $-$ U1b & +4.8\,[$-$4.8, +14.4] & 0.3639 & 1.000 & 0.655 & 1.000 & 0.479 \\
U2-Astra $-$ UCE+P & $-$2.8\,[$-$12.6, +7.0] & 0.6075 & 1.000 & 0.875 & 1.000 & 0.727 \\
U2-Astra $-$ UCE & $-$0.9\,[$-$9.7, +8.0] & 0.8666 & 1.000 & 0.936 & 1.000 & 0.931 \\
U2-Astra $-$ TIME & $-$3.4\,[$-$9.5, +2.1] & 0.2968 & 1.000 & 0.562 & 1.000 & 0.405 \\
RS $-$ U1b & +6.0\,[$-$3.2, +15.3] & 0.2353 & 1.000 & 0.498 & 1.000 & 0.336 \\
RS $-$ UCE+P & $-$1.6\,[$-$11.8, +9.8] & 0.7985 & 1.000 & 0.936 & 1.000 & 0.884 \\
RS $-$ UCE & +0.3\,[$-$9.6, +10.8] & 0.9621 & 1.000 & 0.962 & 1.000 & 0.986 \\
RS $-$ TIME & $-$2.2\,[$-$7.0, +2.3] & 0.4029 & 1.000 & 0.659 & 1.000 & 0.518 \\
U1b $-$ UCE+P & $-$7.6\,[$-$16.4, +2.3] & 0.1405 & 1.000 & 0.361 & 1.000 & 0.230 \\
U1b $-$ UCE & $-$5.6\,[$-$13.0, +2.3] & 0.1791 & 1.000 & 0.403 & 1.000 & 0.269 \\
U1b $-$ TIME & $-$8.2\,[$-$16.1, $-$1.0] & 0.0459 & 1.000 & 0.210 & 1.000 & 0.092 \\
UCE+P $-$ UCE & +1.9\,[$-$1.3, +4.9] & 0.2753 & 1.000 & 0.551 & 1.000 & 0.381 \\
UCE+P $-$ TIME & $-$0.6\,[$-$9.9, +7.1] & 0.9097 & 1.000 & 0.936 & 1.000 & 0.960 \\
UCE $-$ TIME & $-$2.5\,[$-$10.4, +4.2] & 0.5374 & 1.000 & 0.841 & 1.000 & 0.661 \\
\bottomrule
\end{longtable}
\Needspace{12\baselineskip}
\begin{longtable}{p{4.0cm}>{\centering\arraybackslash}p{3.1cm}*{5}{>{\centering\arraybackslash}p{1.55cm}}}

\toprule
contrast & difference [95\% CI] & $p$ & Holm (end.) & BH (end.) & Holm (all) & BH (all) \\
\midrule
\multicolumn{7}{l}{\textit{Table~\ref{tab:A6}, block 7 of 8: native CLIP efficacy}} \\
\endfirsthead
\toprule
\multicolumn{7}{l}{\textit{Table~\ref{tab:A6} (continued): native CLIP efficacy}} \\
contrast & difference [95\% CI] & $p$ & Holm (end.) & BH (end.) & Holm (all) & BH (all) \\
\midrule
\endhead
\midrule\multicolumn{7}{r}{\textit{continued on next page}} \\
\endfoot
\endlastfoot
U3-Opus $-$ U3-Astra & $-$5.2\,[$-$13.9, +0.3] & 0.2195 & 1.000 & 0.272 & 1.000 & 0.316 \\
U3-Opus $-$ U2-Opus & +17.4\,[+3.5, +33.3] & 0.0558 & 0.837 & 0.091 & 1.000 & 0.106 \\
U3-Opus $-$ U2-Astra & $-$4.9\,[$-$14.2, +0.7] & 0.3751 & 1.000 & 0.397 & 1.000 & 0.488 \\
U3-Opus $-$ RS & $-$0.3\,[$-$5.2, +3.8] & 1.0000 & 1.000 & 1.000 & 1.000 & 1.000 \\
U3-Opus $-$ U1b & $-$6.2\,[$-$15.6, $-$0.7] & 0.0788 & 1.000 & 0.118 & 1.000 & 0.139 \\
\textbf{U3-Opus $-$ UCE+P} & +37.2\,[+17.4, +55.6] & 0.0016 & 0.044 & 0.006 & 0.355 & 0.007 \\
\textbf{U3-Opus $-$ UCE} & +30.6\,[+11.8, +48.3] & 0.0044 & 0.105 & 0.012 & 0.873 & 0.014 \\
U3-Opus $-$ TIME & +7.6\,[$-$1.0, +17.4] & 0.1682 & 1.000 & 0.216 & 1.000 & 0.255 \\
\textbf{U3-Astra $-$ U2-Opus} & +22.6\,[+8.0, +39.6] & 0.0154 & 0.308 & 0.033 & 1.000 & 0.040 \\
U3-Astra $-$ U2-Astra & +0.3\,[$-$2.1, +2.4] & 1.0000 & 1.000 & 1.000 & 1.000 & 1.000 \\
U3-Astra $-$ RS & +4.9\,[+1.0, +10.1] & 0.0306 & 0.520 & 0.055 & 1.000 & 0.067 \\
U3-Astra $-$ U1b & $-$1.0\,[$-$2.1, 0.0] & 0.2494 & 1.000 & 0.299 & 1.000 & 0.350 \\
\textbf{U3-Astra $-$ UCE+P} & +42.4\,[+26.7, +58.7] & 0.0001 & 0.002 & $<$0.001 & 0.017 & 0.001 \\
\textbf{U3-Astra $-$ UCE} & +35.8\,[+21.2, +51.0] & $<$0.0001 & 0.001 & $<$0.001 & 0.011 & $<$0.001 \\
\textbf{U3-Astra $-$ TIME} & +12.8\,[+5.2, +21.9] & 0.0024 & 0.060 & 0.007 & 0.496 & 0.008 \\
\textbf{U2-Opus $-$ U2-Astra} & $-$22.2\,[$-$38.9, $-$7.3] & 0.0188 & 0.357 & 0.038 & 1.000 & 0.046 \\
U2-Opus $-$ RS & $-$17.7\,[$-$33.3, $-$4.2] & 0.0358 & 0.572 & 0.061 & 1.000 & 0.075 \\
\textbf{U2-Opus $-$ U1b} & $-$23.6\,[$-$40.3, $-$8.7] & 0.0118 & 0.249 & 0.027 & 1.000 & 0.033 \\
U2-Opus $-$ UCE+P & +19.8\,[$-$5.9, +43.8] & 0.1497 & 1.000 & 0.203 & 1.000 & 0.236 \\
U2-Opus $-$ UCE & +13.2\,[$-$11.5, +36.8] & 0.3130 & 1.000 & 0.352 & 1.000 & 0.423 \\
U2-Opus $-$ TIME & $-$9.7\,[$-$26.4, +5.2] & 0.2748 & 1.000 & 0.319 & 1.000 & 0.381 \\
U2-Astra $-$ RS & +4.5\,[0.0, +10.8] & 0.1524 & 1.000 & 0.203 & 1.000 & 0.237 \\
U2-Astra $-$ U1b & $-$1.4\,[$-$3.8, +0.3] & 0.3743 & 1.000 & 0.397 & 1.000 & 0.488 \\
\textbf{U2-Astra $-$ UCE+P} & +42.0\,[+26.4, +58.0] & $<$0.0001 & 0.001 & $<$0.001 & 0.007 & $<$0.001 \\
\textbf{U2-Astra $-$ UCE} & +35.4\,[+21.2, +50.7] & 0.0001 & 0.004 & 0.001 & 0.029 & 0.001 \\
\textbf{U2-Astra $-$ TIME} & +12.5\,[+5.2, +21.2] & 0.0005 & 0.015 & 0.003 & 0.119 & 0.003 \\
\textbf{RS $-$ U1b} & $-$5.9\,[$-$11.8, $-$1.7] & 0.0076 & 0.167 & 0.018 & 1.000 & 0.022 \\
\textbf{RS $-$ UCE+P} & +37.5\,[+20.5, +54.9] & 0.0006 & 0.018 & 0.003 & 0.143 & 0.003 \\
\textbf{RS $-$ UCE} & +30.9\,[+14.6, +47.6] & 0.0019 & 0.050 & 0.006 & 0.416 & 0.008 \\
RS $-$ TIME & +8.0\,[+1.0, +16.7] & 0.0693 & 0.970 & 0.108 & 1.000 & 0.127 \\
\textbf{U1b $-$ UCE+P} & +43.4\,[+28.1, +59.0] & $<$0.0001 & $<$0.001 & $<$0.001 & 0.001 & $<$0.001 \\
\textbf{U1b $-$ UCE} & +36.8\,[+22.9, +51.7] & $<$0.0001 & 0.001 & $<$0.001 & 0.007 & $<$0.001 \\
\textbf{U1b $-$ TIME} & +13.9\,[+5.9, +23.6] & 0.0008 & 0.023 & 0.003 & 0.183 & 0.004 \\
UCE+P $-$ UCE & $-$6.6\,[$-$15.3, 0.0] & 0.1472 & 1.000 & 0.203 & 1.000 & 0.234 \\
\textbf{UCE+P $-$ TIME} & $-$29.5\,[$-$46.2, $-$11.8] & 0.0046 & 0.105 & 0.012 & 0.903 & 0.014 \\
UCE $-$ TIME & $-$22.9\,[$-$39.9, $-$5.2] & 0.0231 & 0.415 & 0.044 & 1.000 & 0.054 \\
\bottomrule
\end{longtable}
\Needspace{12\baselineskip}
\begin{longtable}{p{4.0cm}>{\centering\arraybackslash}p{3.1cm}*{5}{>{\centering\arraybackslash}p{1.55cm}}}

\toprule
contrast & difference [95\% CI] & $p$ & Holm (end.) & BH (end.) & Holm (all) & BH (all) \\
\midrule
\multicolumn{7}{l}{\textit{Table~\ref{tab:A6}, block 8 of 8: native CLIP generality}} \\
\endfirsthead
\toprule
\multicolumn{7}{l}{\textit{Table~\ref{tab:A6} (continued): native CLIP generality}} \\
contrast & difference [95\% CI] & $p$ & Holm (end.) & BH (end.) & Holm (all) & BH (all) \\
\midrule
\endhead
\midrule\multicolumn{7}{r}{\textit{continued on next page}} \\
\endfoot
\endlastfoot
U3-Opus $-$ U3-Astra & $-$1.6\,[$-$7.1, +3.8] & 0.6024 & 1.000 & 0.620 & 1.000 & 0.726 \\
\textbf{U3-Opus $-$ U2-Opus} & +20.1\,[+8.4, +33.3] & 0.0023 & 0.042 & 0.004 & 0.477 & 0.008 \\
U3-Opus $-$ U2-Astra & +1.9\,[$-$4.2, +7.9] & 0.5587 & 1.000 & 0.592 & 1.000 & 0.685 \\
\textbf{U3-Opus $-$ RS} & +7.2\,[+2.3, +13.6] & 0.0081 & 0.121 & 0.013 & 1.000 & 0.024 \\
U3-Opus $-$ U1b & $-$3.5\,[$-$10.0, +2.4] & 0.3147 & 1.000 & 0.354 & 1.000 & 0.424 \\
\textbf{U3-Opus $-$ UCE+P} & +46.5\,[+30.2, +61.5] & $<$0.0001 & 0.001 & $<$0.001 & 0.008 & $<$0.001 \\
\textbf{U3-Opus $-$ UCE} & +37.8\,[+22.0, +52.7] & 0.0002 & 0.005 & 0.001 & 0.053 & 0.001 \\
\textbf{U3-Opus $-$ TIME} & +23.1\,[+12.3, +34.0] & 0.0004 & 0.009 & 0.001 & 0.092 & 0.002 \\
\textbf{U3-Astra $-$ U2-Opus} & +21.7\,[+10.6, +34.4] & 0.0004 & 0.009 & 0.001 & 0.094 & 0.002 \\
U3-Astra $-$ U2-Astra & +3.5\,[$-$0.7, +8.2] & 0.1511 & 0.907 & 0.176 & 1.000 & 0.237 \\
\textbf{U3-Astra $-$ RS} & +8.8\,[+2.6, +15.8] & 0.0122 & 0.158 & 0.018 & 1.000 & 0.033 \\
U3-Astra $-$ U1b & $-$1.9\,[$-$6.9, +2.9] & 0.4928 & 1.000 & 0.538 & 1.000 & 0.623 \\
\textbf{U3-Astra $-$ UCE+P} & +48.1\,[+34.0, +61.6] & $<$0.0001 & $<$0.001 & $<$0.001 & 0.001 & $<$0.001 \\
\textbf{U3-Astra $-$ UCE} & +39.4\,[+25.3, +52.9] & 0.0001 & 0.002 & $<$0.001 & 0.017 & 0.001 \\
\textbf{U3-Astra $-$ TIME} & +24.7\,[+15.1, +34.8] & 0.0001 & 0.002 & $<$0.001 & 0.014 & 0.001 \\
\textbf{U2-Opus $-$ U2-Astra} & $-$18.2\,[$-$31.7, $-$6.2] & 0.0089 & 0.125 & 0.014 & 1.000 & 0.026 \\
U2-Opus $-$ RS & $-$12.9\,[$-$25.8, $-$1.7] & 0.0503 & 0.503 & 0.067 & 1.000 & 0.098 \\
\textbf{U2-Opus $-$ U1b} & $-$23.6\,[$-$38.1, $-$10.3] & 0.0019 & 0.037 & 0.004 & 0.405 & 0.007 \\
\textbf{U2-Opus $-$ UCE+P} & +26.4\,[+7.6, +44.1] & 0.0127 & 0.158 & 0.018 & 1.000 & 0.034 \\
U2-Opus $-$ UCE & +17.6\,[$-$1.0, +35.3] & 0.0791 & 0.632 & 0.098 & 1.000 & 0.139 \\
U2-Opus $-$ TIME & +3.0\,[$-$12.2, +16.3] & 0.7005 & 1.000 & 0.701 & 1.000 & 0.807 \\
U2-Astra $-$ RS & +5.3\,[$-$0.8, +11.4] & 0.1121 & 0.785 & 0.135 & 1.000 & 0.189 \\
U2-Astra $-$ U1b & $-$5.4\,[$-$9.9, $-$1.0] & 0.0298 & 0.327 & 0.041 & 1.000 & 0.067 \\
\textbf{U2-Astra $-$ UCE+P} & +44.6\,[+31.5, +57.3] & $<$0.0001 & 0.001 & $<$0.001 & 0.006 & $<$0.001 \\
\textbf{U2-Astra $-$ UCE} & +35.8\,[+22.7, +48.8] & $<$0.0001 & 0.001 & $<$0.001 & 0.012 & $<$0.001 \\
\textbf{U2-Astra $-$ TIME} & +21.2\,[+12.2, +30.6] & 0.0002 & 0.005 & 0.001 & 0.050 & 0.001 \\
\textbf{RS $-$ U1b} & $-$10.7\,[$-$18.3, $-$3.8] & 0.0074 & 0.118 & 0.013 & 1.000 & 0.022 \\
\textbf{RS $-$ UCE+P} & +39.3\,[+23.1, +54.6] & 0.0002 & 0.005 & 0.001 & 0.053 & 0.001 \\
\textbf{RS $-$ UCE} & +30.6\,[+14.2, +46.5] & 0.0022 & 0.042 & 0.004 & 0.469 & 0.008 \\
\textbf{RS $-$ TIME} & +15.9\,[+7.4, +24.3] & 0.0017 & 0.036 & 0.004 & 0.374 & 0.007 \\
\textbf{U1b $-$ UCE+P} & +50.0\,[+38.1, +61.2] & $<$0.0001 & $<$0.001 & $<$0.001 & 0.003 & $<$0.001 \\
\textbf{U1b $-$ UCE} & +41.3\,[+29.3, +52.8] & $<$0.0001 & $<$0.001 & $<$0.001 & 0.004 & $<$0.001 \\
\textbf{U1b $-$ TIME} & +26.6\,[+18.0, +36.0] & $<$0.0001 & $<$0.001 & $<$0.001 & 0.001 & $<$0.001 \\
\textbf{UCE+P $-$ UCE} & $-$8.7\,[$-$12.1, $-$5.6] & $<$0.0001 & $<$0.001 & $<$0.001 & 0.001 & $<$0.001 \\
\textbf{UCE+P $-$ TIME} & $-$23.4\,[$-$36.0, $-$8.8] & 0.0042 & 0.071 & 0.007 & 0.844 & 0.014 \\
UCE $-$ TIME & $-$14.7\,[$-$28.1, +0.2] & 0.0611 & 0.550 & 0.079 & 1.000 & 0.114 \\
\bottomrule
\end{longtable}
}

\end{document}